%% file: root.tex
\documentclass[lettersize,journal]{IEEEtran}

\input{packages}
\begin{document}

\title{\LARGE \bf Cooperative Energy and Time-Optimal Lane Change Maneuvers with Minimal Highway Traffic Disruption 
\thanks{Supported by the Honda Research Institute USA (HRI-USA), by NSF under grants ECCS-1931600,
DMS-1664644, CNS-1645681, CNS-2149511, by AFOSR under grant FA9550-19-1-0158,
by ARPA-E under grant DE-AR0001282, and by the MathWorks}
}

\author{Andres S. Chavez Armijos$^{1}$, Anni Li$^{1}$, Christos G. Cassandras$^{1}$,
\thanks{$^{1}$A. S. Chavez Armijos, A. Li, and C. G. Cassandras are
with the Division of Systems Engineering and Center for Information and
Systems Engineering, Boston University, Brookline, MA 02446
(email:\{aschavez; anlianni; cgc\}@bu.edu).}
\\Yasir K. Al-Nadawi$^{2}$, Hidekazu Araki$^{2}$, Behdad Chalaki$^2$, Ehsan Moradi-Pari$^{2}$, \\Hossein Nourkhiz Mahjoub$^{2}$, and Vaishnav Tadiparthi$^2$
 
\thanks{$^{2}$H. Nourkhiz Mahjoub, B. Chalaki, V. Tadiparthi, E. Moradi-Pari, and H. Araki are with Honda Research Institute-US (HRI-US)
Ann Arbor, MI 48103 USA
(email:\{hnourkhizmahjoub; behdad\_chalaki; vaishnav\_tadiparthi; emoradipari; haraki\}@honda-ri.com) Y. K. Al-Nadawi was with HRI during the period of this work.}
}

\maketitle
\begin{abstract}
\subfile{sections/abstract}
\end{abstract}

\section{Introduction}
\subfile{sections/introduction}
\section{Problem Formulation}\label{secII:ProblemFormulation}
\subfile{sections/problem_formulation}

\section{Longitudinal Optimal Control Solution}\label{secIII:DecentralizedOCSol}
\subfile{sections/long_optimal_control_solution}

\section{Lateral Optimal Control Solution}\label{secIV:LatSol}
\subfile{sections/lat_optimal_control_solution}
\section{Sequential Maneuvers}\label{secV:Sequential_Maneuvers}
\subfile{sections/sequential_maneuvers}
\section{Simulation Results}\label{secVI:ExperimentalSolution}
\subfile{sections/simulation_results}

\section{Conclusions}\label{secVII:Conclusions}
\subfile{sections/Conclusions}

\appendix
\subfile{sections/appendix}

\bibliographystyle{IEEEtran}
\begin{tiny}
\bibliography{bibliography,cmp}
\end{tiny}

\end{document}

%% file: packages.tex
\usepackage[OT1]{fontenc} 
\usepackage[english]{babel}
\usepackage{cite}
\usepackage{amsmath,amssymb,amsfonts}
\usepackage{textcomp}
\usepackage{xcolor}
\usepackage{blindtext}
\usepackage{subfiles} 
\usepackage{mathtools}
\usepackage{physics}
\usepackage{tikz}
\usepackage{svg}
\usepackage{booktabs} 
\usepackage[final]{microtype} 
\usepackage{hyperref}
\hypersetup{hidelinks, allcolors=black}
\usepackage[capitalise]{cleveref}
\usepackage{mathdots}
\usepackage{yhmath}
\usepackage{cancel}
\usepackage{color}
\usepackage{siunitx}
\usepackage{array}
\usepackage{multirow}
\usepackage{amssymb}
\usepackage{gensymb}
\usepackage{adjustbox}
\usepackage{tabularx}
\usepackage{comment}
\usepackage{extarrows}
\usepackage{booktabs}
\usepackage{diagbox}
\usepackage{graphicx}
\usepackage{subcaption}
\usepackage[normalem]{ulem}
\useunder{\uline}{\ul}{}
\usetikzlibrary{fadings}
\usetikzlibrary{patterns}
\usetikzlibrary{shadows.blur}
\usetikzlibrary{shapes}
\graphicspath{ {./Figures/} }

\DeclareMathOperator*{\argmin}{argmin}

\newtheorem{rem}{Remark}

\def\BibTeX{{\rm B\kern-.05em{\sc i\kern-.025em b}\kern-.08em
    T\kern-.1667em\lower.7ex\hbox{E}\kern-.125emX}}

%% file: sections/abstract.tex
We derive optimal control policies for a Connected Automated Vehicle (CAV) and cooperating neighboring CAVs to carry out a lane change maneuver consisting of a longitudinal phase where the CAV properly positions itself relative to the cooperating neighbors and a lateral phase where
it safely changes lanes. 
In contrast to prior work on this problem, where the CAV ``selfishly'' only seeks to minimize its maneuver time, we seek to ensure that the fast-lane traffic flow is minimally disrupted (through a properly defined metric). Additionally, when performing lane-changing maneuvers, we optimally select the cooperating vehicles from a set of feasible neighboring vehicles and experimentally show that the highway throughput is improved compared to the baseline case of human-driven vehicles changing lanes with no cooperation. 
When feasible solutions do not exist for a given maximal allowable disruption, we include a time relaxation method trading off a longer maneuver time with reduced disruption. Our analysis is also extended to multiple sequential maneuvers. Simulation results show the effectiveness of our controllers in terms of safety guarantees and up to 16\% and 90\% average throughput and maneuver time improvement respectively when compared to maneuvers with no vehicle cooperation.

%% file: sections/introduction.tex
Advances in transportation system technologies and the emergence of Connected
Automated Vehicles (CAVs), also known as \textquotedblleft autonomous
vehicles\textquotedblright, have the potential to drastically improve a
transportation network's performance in terms of safety, comfort, congestion
reduction and energy efficiency. 

In highway driving, an overview of automated
intelligent vehicle-highway systems was provided in \cite{varaiya1993smart}. Recent developments focusing on autonomous car-following
control are given in \cite{zhao2018accelerated},\cite{wang2016cooperative}%
,\cite{wang2015game}. However, automating a lane change maneuver remains a challenging problem that has
attracted increasing attention in recent years \cite{nilsson2015longitudinal, bax2014road, he2021rule}. Work to date is restricted to controlling a single vehicle during the maneuver and no analysis of the overall disruption effect that various proposed solutions produce on the traffic flow has been provided.

The emergence of CAVs creates the opportunity for cooperation among vehicles
traveling on multi-lane roads to carry out automated
lane-change maneuvers \cite{mahjoub2017learning}, \cite{luo2016dynamic}, \cite{li2020cooperative}. In particular, when
controlling a single vehicle and checking on the feasibility of a maneuver
depending on the state of nearby traffic, as in \cite{kamal2013model}%
,\cite{katriniok2013optimal}, the maneuver is often infeasible without the
cooperation of other vehicles, especially under heavier traffic conditions. In
contrast, a cooperative approach can allow multiple interacting vehicles to
implement controllers enabling a larger set of feasible maneuvers. Aside from enhancing
safety, this cooperative behavior can also improve the throughput, hence reducing the chance of congestion. 

The problem of cooperative multi-agent lane-changing maneuvers can be solved as a centralized or decentralized motion planning problem. In the centralized case, a Control Zone (CZ) is defined as a prespecified area within which all vehicles follow commands issued by a central coordinator (e.g. traffic beacon) that computes a solution for every vehicle involved. This makes the computation intractable on some occasions. In contrast, in the decentralized case, the multi-agent problem is solved by each individual agent computing their individual solution on board. In this case, solutions may be too conservative and could generate unwanted disruptions in traffic  \cite{li2018balancing}.

Feasible, but not necessarily optimal,
vehicle trajectories for cooperative multi-agent lane-changing maneuvers are
derived in \cite{lam2013cooperative}. The case of multiple cooperating
vehicles simultaneously changing lanes is considered in \cite{li2017optimal}
with the requirement that all vehicles are controllable and their velocities
before the lane change are all the same. First, vehicles with a lower
priority must adjust their positions in their current lane and give way to
those with a higher priority to avoid collisions. Then, an optimal control problem (OCP) is solved for each vehicle without considering the
usual safe distance constraints between vehicles. This \textquotedblleft
progressively constrained dynamic optimization\textquotedblright\ method
facilitates a numerical solution to the underlying OCP at
the expense of loss in performance.

\begin{figure} [pt]
    \centering
    \begin{adjustbox}{width=7cm, height = 3cm,center}
    \begin{tikzpicture}[x=0.75pt,y=0.75pt,yscale=-0.55,xscale=0.44]
        \draw [line width=3]    (0,35.16) -- (358.76,35.7) -- (740,35.7) ;
        \draw [line width=3]    (0,239.03) -- (321.62,239.03) -- (740,239.03) ;
        \draw [color={rgb, 255:red, 248; green, 231; blue, 28 }  ,draw opacity=1 ][fill={rgb, 255:red, 248; green, 231; blue, 28 }  ,fill opacity=1 ][line width=3]  [dash pattern={on 13pt off 10pt}]  (0,138.11) -- (740.81,136.69) ;
        \draw (500,185) node  {\includegraphics[scale=0.11]{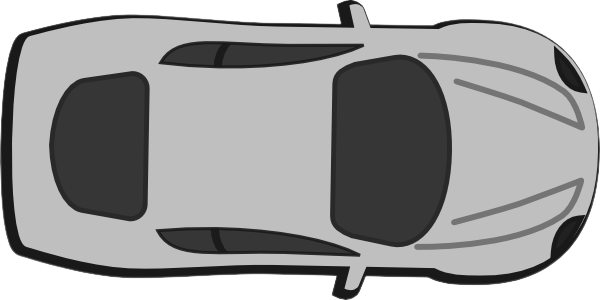}};
        \draw (212.09,185) node  {\includegraphics[scale=0.07]{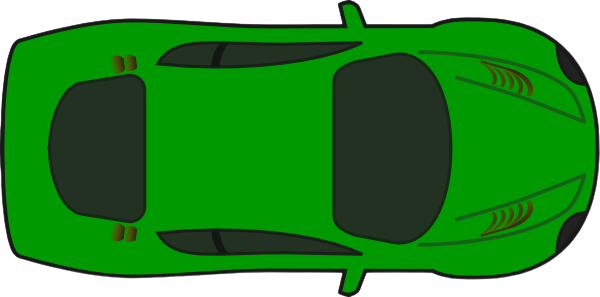}};
        \draw (514.97,90) node  {\includegraphics[scale=0.17]{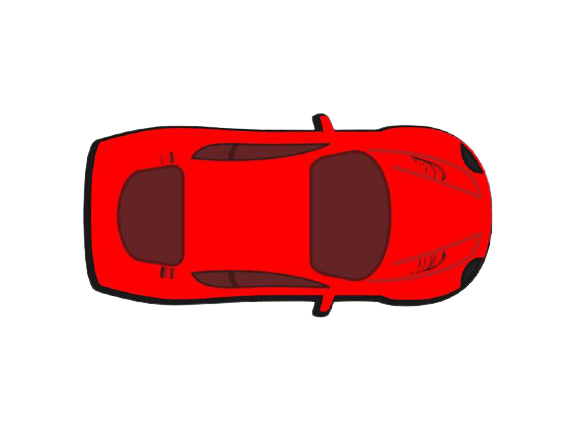}};
        \draw (195.97,90) node  {\includegraphics[scale=0.17]{redVehicleTopView.png}};
        \draw [color={rgb, 255:red, 65; green, 117; blue, 5 }  ,draw opacity=1 ] [dash pattern={on 3.75pt off 3pt on 7.5pt off 1.5pt}]  (270.18,184.81) .. controls (368.48,192.9) and (296.46,68) .. (417.56,80.51) ;
        \draw [shift={(419.4,80.71)}, rotate = 186.44] [fill={rgb, 255:red, 65; green, 117; blue, 5 }  ,fill opacity=1 ][line width=0.08]  [draw opacity=0] (10.72,-5.15) -- (0,0) -- (10.72,5.15) -- (7.12,0) -- cycle    ;
        \draw [color={rgb, 255:red, 208; green, 2; blue, 27 }  ,draw opacity=1 ][fill={rgb, 255:red, 208; green, 2; blue, 27 }  ,fill opacity=1 ] [dash pattern={on 3.75pt off 3pt on 7.5pt off 1.5pt}]  (114.82,90) -- (81.38,90) ;
        \draw [shift={(79.38,90)}, rotate = 360] [color={rgb, 255:red, 208; green, 2; blue, 27 }  ,draw opacity=1 ][line width=0.75]    (10.93,-3.29) .. controls (6.95,-1.4) and (3.31,-0.3) .. (0,0) .. controls (3.31,0.3) and (6.95,1.4) .. (10.93,3.29)   ;
        \draw [color={rgb, 255:red, 208; green, 2; blue, 27 }  ,draw opacity=1 ] [dash pattern={on 3.75pt off 3pt on 7.5pt off 1.5pt}]  (576.12,90) -- (620.51,90) ;
        \draw [shift={(620.51,90)}, rotate = 180] [color={rgb, 255:red, 208; green, 2; blue, 27 }  ,draw opacity=1 ][line width=0.75]    (10.93,-3.29) .. controls (6.95,-1.4) and (3.31,-0.3) .. (0,0) .. controls (3.31,0.3) and (6.95,1.4) .. (10.93,3.29)   ;
        
        \draw (0,45) node [anchor=north west][inner sep=0.3pt]  [font=\large] [align=left] {$\displaystyle u_{2}^{*} \leq 0$$ $};
        \draw (592.81,48.15) node [anchor=north west][inner sep=0.75pt]  [font=\large] [align=left] {$\displaystyle u_{1}^{*} \geq 0$$ $};
        \draw (335.63,145.53) node [anchor=north west][inner sep=0.75pt]  [font=\large] [align=left] {$\displaystyle u_{C}^{*}$$ $};
        \draw (215.95,183) node  [font=\large] [align=left] {\begin{minipage}[lt]{26.64pt}\setlength\topsep{0pt}
        C
        \end{minipage}};
        \draw (510.26,183) node  [font=\large] [align=left] {\begin{minipage}[lt]{26.64pt}\setlength\topsep{0pt}
        U
        \end{minipage}};
        \draw (525.44,90) node  [font=\large] [align=left] {\begin{minipage}[lt]{26.64pt}\setlength\topsep{0pt}
        1
        \end{minipage}};
        \draw (210,90) node  [font=\large] [align=left] {\begin{minipage}[lt]{26.64pt}\setlength\topsep{0pt}
        2
        \end{minipage}};
    \end{tikzpicture}
    \end{adjustbox}
    \caption{The basic lane-changing maneuver process.}
    \label{Fig1:original_maneuver_diagram}
    \vspace*{-\baselineskip}
    \vspace*{-3mm}
\end{figure}

In previous work, \cite{chen2020cooperative} provided a time and energy optimal solution for the maneuver shown in
Fig. \ref{Fig1:original_maneuver_diagram}, in which the controlled vehicle $C$ attempts to
overtake an uncontrollable vehicle $U$ by using the left lane to pass. 
A decentralized solution is provided based on cooperation and communication with two neighboring vehicles (vehicles $1$ and $2$) with the goal of minimizing the total maneuver time and subsequently determining trajectories that minimize the energy consumed by all three cooperating vehicles. 
This approach applies to a wider range of scenarios
relative to those in \cite{nilsson2017lane, luo2016dynamic, kamal2013model}. 

However, by seeking to minimize $C$'s maneuver time, \cite{chen2020cooperative} adopts a \emph{vehicle-centric} (selfish) viewpoint which ignores the effect of the maneuver on all remaining vehicles. As a result, since vehicle $2$ typically decelerates to allow $C$ to get ahead of it, this deceleration may cause a traffic flow slowdown in the left lane which can negatively impact throughput, especially in congested scenarios. Moreover, the analysis assumes that vehicles $1$ and $2$ are predetermined rather than being optimally selected among a set of possible cooperation candidates.

The main contribution of this paper is the alleviation of the aforementioned limitations in such \emph{selfish} maneuvers by adopting an optimality viewpoint that combines 
\emph{system-centric} objectives with vehicle-centric ones, as introduced in \cite{armijos2022sequential}. This provides a decentralized optimal solution where our optimal controller design includes the social optimality goal of ensuring that the resulting traffic throughput on the highway is improved by adding vehicles to the fast lane while (i) limiting the ``disruption'' that cooperation among multiple vehicles on the road can cause on the fast lane traffic flow, and (ii) determining an optimal pair of cooperating vehicles, which play the role of $1$ and $2$ in Fig. \ref{Fig1:original_maneuver_diagram}, within a set of feasible such candidates. Additionally, we focus on the realization of \emph{multiple} sequential lane-changing maneuvers under the assumption of cooperation with surrounding vehicles. 

The disruption metric introduced in \cite{armijos2022sequential} considers how the \emph{positions} of vehicles in the fast lane are disrupted relative to the ``ideal'' trajectories where fixed speeds are maintained. The associated disruption in their \emph{speeds} is, however, not taken into account and this can be shown to allow for significant slowdowns in the traffic flow in certain situations. A key contribution of this paper is to alleviate this limitation in \cite{armijos2022sequential} by employing a disruption metric that includes both vehicle position and speed.
This necessitates a new analysis for the problem of deriving time and energy optimal maneuvers while also ensuring throughput improvements under different traffic densities. This paper provides this analysis, considers the sensitivity of the resulting algorithm to the aforementioned traffic densities, and makes use of a new estimate of the traffic free flow speed based on the speeds from the cooperative vehicle sets. 

As in \cite{chen2020cooperative}, we decompose the maneuver into a longitudinal component followed by a lateral component. In the longitudinal part, our approach is based on first determining an optimal maneuver time for $C$ subject to all safety and speed and acceleration constraints for $C$,
$1$, and $2$ (see Fig. \ref{Fig1:original_maneuver_diagram}) and such that $C$ attains a desired final speed that matches that of the fast lane traffic flow. We then solve a fixed terminal time decentralized optimal control
problem for each of the two cooperating vehicles in which energy
consumption is minimized while penalizing the deviation of $1$ and $2$ from the flow speed. In the lateral phase, we solve a decentralized optimal control
problem seeking to jointly minimize the time and energy consumed which is no different than the one presented in \cite{chen2020cooperative}.
Our analysis also allows the determination of a vehicle pair that results in minimal acceleration/deceleration for them. This minimizes the possibility of excessive braking or deceleration of the rear vehicle in the pair ($2$ in Fig. \ref{Fig1:original_maneuver_diagram}), quantified through an appropriate ``disruption metric''.
An interesting consequence of our analysis is that it leads to vehicles forming natural platoons that dictate the free-flow speed of the fast lane on a two-lane highway. 

The rest of the paper is organized as follows. Section \ref{secII:ProblemFormulation} presents the formulation of the
longitudinal lane-change maneuver problem. In Section \ref{secIII:DecentralizedOCSol},
a complete optimal control solution to coordinate the longitudinal portion of the lane change maneuver is obtained. Section \ref{secIV:LatSol} describes the lateral portion of a lane-changing maneuver. Section \ref{secV:Sequential_Maneuvers} deals with sequential multiple maneuvers.
Section \ref{secVI:ExperimentalSolution} provides simulation results for
several representative examples and we conclude with Section \ref{secVII:Conclusions}.

%% file: sections/problem_formulation.tex
In this section, we present a system-centric problem formulation of the cooperative maneuver setting in Fig. \ref{Fig1:original_maneuver_diagram}. We decompose the maneuver into a longitudinal component followed by a lateral component. The former includes the determination of a minimally disrupting cooperative pair (playing the role of $1$ and $2$ in Fig. \ref{Fig1:original_maneuver_diagram}) while minimizing the deviation from the fast lane flow speed. 

Let $C$ be the vehicle that initiates an automated maneuver. This can be manually triggered by the driver of $C$ deciding to overtake vehicle $U$
or automatically triggered by a given distance detected from an uncontrollable vehicle $U$ ahead of $C$, as shown in Fig. \ref{fig3:cav_set_selection}. Assuming that all vehicles other than $U$ are CAVs, we will henceforth refer to them as such.

\subsection{Cooperative Vehicle Set}
\begin{figure}[pt]
    \centering
    \vspace*{1mm}
    \begin{adjustbox}{width=\linewidth, height = 4cm,center}
    
    \tikzset{every picture/.style={line width=0.75pt}} 
    
    \begin{tikzpicture}[x=0.75pt,y=0.75pt,yscale=-1,xscale=1]
    
        \draw [line width=3]    (50.81,39.52) -- (239.1,39.21) -- (445.37,39.35) ;
        \draw [line width=3]    (50.43,130.42) -- (218.61,131.02) -- (444.99,130.59) ;
        \draw [color={rgb, 255:red, 248; green, 231; blue, 28 }  ,draw opacity=1 ][fill={rgb, 255:red, 248; green, 231; blue, 28 }  ,fill opacity=1 ][line width=3]  [dash pattern={on 11.25pt off 9.75pt}]  (51.33,89.28) -- (446,88.77) ;
        \draw (343.84,104.67) node [rotate=-0.02,xslant=0] {\includegraphics[width=26.23pt,height=16.6pt]{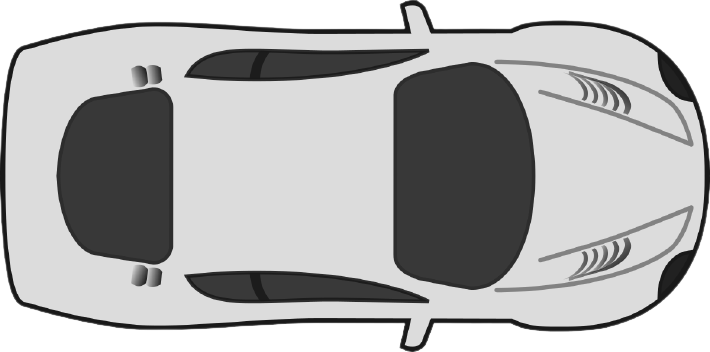}};
        \draw (284.48,59.06) node [rotate=-0.02,xslant=0] {\includegraphics[width=26.23pt,height=16.6pt]{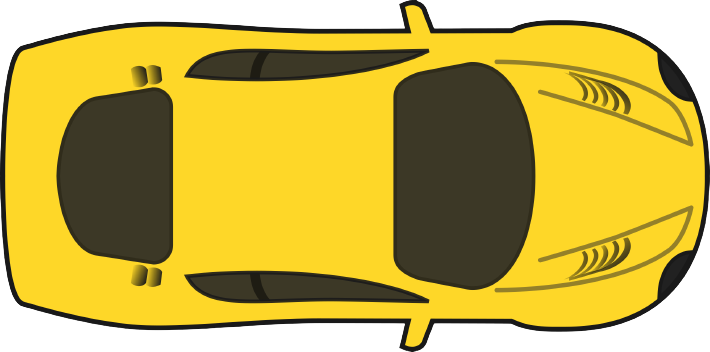}};
        \draw (421.59,59.13) node [rotate=-0.02,xslant=0] {\includegraphics[width=26.23pt,height=16.6pt]{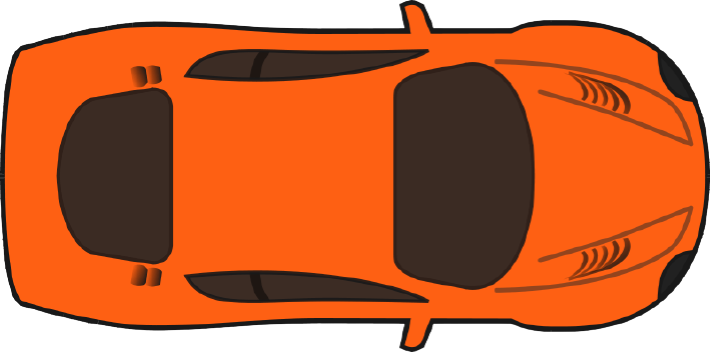}};
        \draw (361.04,59.1) node [rotate=-0.02,xslant=0] {\includegraphics[width=26.23pt,height=16.6pt]{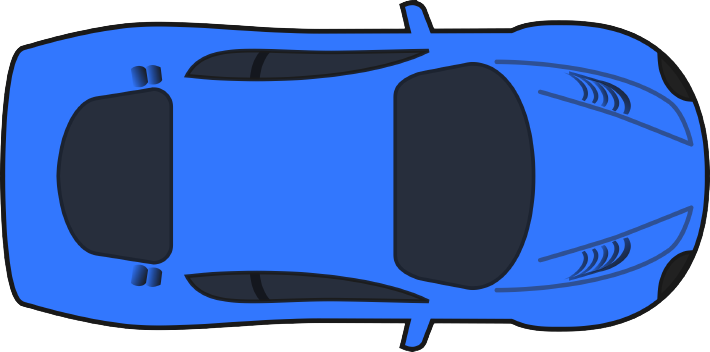}};
        \draw (260.03,104.17) node [rotate=-0.02,xslant=0] {\includegraphics[width=26.23pt,height=16.6pt]{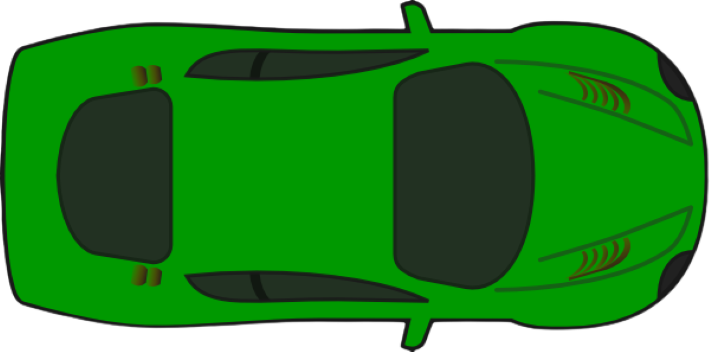}};
        \draw (214.3,58.57) node [rotate=-0.02,xslant=0] {\includegraphics[width=26.23pt,height=16.6pt]{car_yellow.png}};
        \draw (144.84,59.02) node [rotate=-0.02,xslant=0] {\includegraphics[width=26.23pt,height=16.6pt]{car_blue.png}};
        \draw (75.41,58.96) node [rotate=-0.02,xslant=0] {\includegraphics[width=26.23pt,height=16.6pt]{car_orange.png}};
        \draw (156.21,105.55) node [rotate=-0.02,xslant=0] {\includegraphics[width=26.23pt,height=16.6pt]{car_orange.png}};
        \draw  [color={rgb, 255:red, 208; green, 2; blue, 27 }  ,draw opacity=1 ][dash pattern={on 4.5pt off 4.5pt}] (111.4,51.5) .. controls (111.41,46.8) and (115.22,42.99) .. (119.92,42.99) -- (386.18,43.09) .. controls (390.88,43.09) and (394.69,46.9) .. (394.68,51.6) -- (394.66,77.11) .. controls (394.65,81.81) and (390.84,85.62) .. (386.14,85.62) -- (119.87,85.53) .. controls (115.18,85.52) and (111.37,81.71) .. (111.38,77.02) -- cycle ;
        \draw [color={rgb, 255:red, 155; green, 155; blue, 155 }  ,draw opacity=1 ][fill={rgb, 255:red, 155; green, 155; blue, 155 }  ,fill opacity=1 ] [dash pattern={on 3.75pt off 3pt on 7.5pt off 1.5pt}]  (239.09,103.67) -- (239,145.95) ;
        \draw [color={rgb, 255:red, 155; green, 155; blue, 155 }  ,draw opacity=1 ][fill={rgb, 255:red, 155; green, 155; blue, 155 }  ,fill opacity=1 ] [dash pattern={on 3.75pt off 3pt on 7.5pt off 1.5pt}]  (175.18,145.92) -- (113.32,145.89) ;
        \draw [shift={(111.32,145.89)}, rotate = 0.02] [color={rgb, 255:red, 155; green, 155; blue, 155 }  ,draw opacity=1 ][line width=0.75]    (10.93,-3.29) .. controls (6.95,-1.4) and (3.31,-0.3) .. (0,0) .. controls (3.31,0.3) and (6.95,1.4) .. (10.93,3.29)   ;
        \draw [color={rgb, 255:red, 155; green, 155; blue, 155 }  ,draw opacity=1 ][fill={rgb, 255:red, 155; green, 155; blue, 155 }  ,fill opacity=1 ] [dash pattern={on 3.75pt off 3pt on 7.5pt off 1.5pt}]  (175.25,145.92) -- (237.47,145.95) ;
        \draw [shift={(239.47,145.95)}, rotate = 180.02] [color={rgb, 255:red, 155; green, 155; blue, 155 }  ,draw opacity=1 ][line width=0.75]    (10.93,-3.29) .. controls (6.95,-1.4) and (3.31,-0.3) .. (0,0) .. controls (3.31,0.3) and (6.95,1.4) .. (10.93,3.29)   ;
        \draw [color={rgb, 255:red, 155; green, 155; blue, 155 }  ,draw opacity=1 ][fill={rgb, 255:red, 155; green, 155; blue, 155 }  ,fill opacity=1 ] [dash pattern={on 3.75pt off 3pt on 7.5pt off 1.5pt}]  (111.39,62.19) -- (111.32,145.89) ;
        
        \draw [color={rgb, 255:red, 155; green, 155; blue, 155 }  ,draw opacity=1 ] [dash pattern={on 3.75pt off 3pt on 7.5pt off 1.5pt}]  (368.31,104.71) -- (368.29,146.25) ;
        \draw [color={rgb, 255:red, 155; green, 155; blue, 155 }  ,draw opacity=1 ] [dash pattern={on 3.75pt off 3pt on 7.5pt off 1.5pt}]  (381.44,146.26) -- (392.6,146.26) ;
        \draw [shift={(394.6,146.26)}, rotate = 180.02] [color={rgb, 255:red, 155; green, 155; blue, 155 }  ,draw opacity=1 ][line width=0.75]    (10.93,-3.29) .. controls (6.95,-1.4) and (3.31,-0.3) .. (0,0) .. controls (3.31,0.3) and (6.95,1.4) .. (10.93,3.29)   ;
        \draw [color={rgb, 255:red, 155; green, 155; blue, 155 }  ,draw opacity=1 ] [dash pattern={on 3.75pt off 3pt on 7.5pt off 1.5pt}]  (381.42,146.26) -- (370.19,146.25) ;
        \draw [shift={(368.19,146.25)}, rotate = 0.02] [color={rgb, 255:red, 155; green, 155; blue, 155 }  ,draw opacity=1 ][line width=0.75]    (10.93,-3.29) .. controls (6.95,-1.4) and (3.31,-0.3) .. (0,0) .. controls (3.31,0.3) and (6.95,1.4) .. (10.93,3.29)   ;
        \draw [color={rgb, 255:red, 155; green, 155; blue, 155 }  ,draw opacity=1 ] [dash pattern={on 3.75pt off 3pt on 7.5pt off 1.5pt}]  (394.67,64.01) -- (394.6,146.26) ;
        
        \draw  [dash pattern={on 0.75pt off 0.75pt}]  (278.37,103.85) .. controls (280.06,102.2) and (281.72,102.22) .. (283.37,103.91) -- (287.48,103.97) -- (295.48,104.07) ;
        \draw [shift={(297.48,104.09)}, rotate = 180.73] [color={rgb, 255:red, 0; green, 0; blue, 0 }  ][line width=0.75]    (10.93,-3.29) .. controls (6.95,-1.4) and (3.31,-0.3) .. (0,0) .. controls (3.31,0.3) and (6.95,1.4) .. (10.93,3.29)   ;
        \draw  [color={rgb, 255:red, 245; green, 166; blue, 35 }  ,draw opacity=1 ][dash pattern={on 4.5pt off 4.5pt}] (184.09,52.4) .. controls (184.1,48.16) and (187.53,44.73) .. (191.77,44.73) -- (303.01,44.77) .. controls (307.24,44.77) and (310.68,48.21) .. (310.67,52.44) -- (310.65,75.45) .. controls (310.64,79.69) and (307.2,83.12) .. (302.97,83.12) -- (191.73,83.08) .. controls (187.49,83.08) and (184.06,79.64) .. (184.07,75.41) -- cycle ;
        \draw [color={rgb, 255:red, 245; green, 166; blue, 35 }  ,draw opacity=1 ]   (288.44,44.67) -- (288.88,28.7) ;
        \draw [shift={(288.93,26.7)}, rotate = 91.57] [color={rgb, 255:red, 245; green, 166; blue, 35 }  ,draw opacity=1 ][line width=0.75]    (10.93,-3.29) .. controls (6.95,-1.4) and (3.31,-0.3) .. (0,0) .. controls (3.31,0.3) and (6.95,1.4) .. (10.93,3.29)   ;
        \draw [color={rgb, 255:red, 208; green, 2; blue, 27 }  ,draw opacity=1 ]   (377.8,43.46) -- (377.43,27.95) ;
        \draw [shift={(377.38,25.95)}, rotate = 88.61] [color={rgb, 255:red, 208; green, 2; blue, 27 }  ,draw opacity=1 ][line width=0.75]    (10.93,-3.29) .. controls (6.95,-1.4) and (3.31,-0.3) .. (0,0) .. controls (3.31,0.3) and (6.95,1.4) .. (10.93,3.29)   ;
        \draw  [color={rgb, 255:red, 65; green, 117; blue, 5 }  ,draw opacity=1 ][dash pattern={on 3.75pt off 3pt on 7.5pt off 1.5pt}] (52.6,51.22) .. controls (52.6,46.35) and (56.55,42.4) .. (61.42,42.4) -- (434.58,42.4) .. controls (439.45,42.4) and (443.4,46.35) .. (443.4,51.22) -- (443.4,77.68) .. controls (443.4,82.55) and (439.45,86.5) .. (434.58,86.5) -- (61.42,86.5) .. controls (56.55,86.5) and (52.6,82.55) .. (52.6,77.68) -- cycle ;
        \draw [color={rgb, 255:red, 65; green, 117; blue, 5 }  ,draw opacity=1 ]   (110.67,42.73) -- (110.07,25.73) ;
        \draw [shift={(110,23.73)}, rotate = 87.99] [color={rgb, 255:red, 65; green, 117; blue, 5 }  ,draw opacity=1 ][line width=0.75]    (10.93,-3.29) .. controls (6.95,-1.4) and (3.31,-0.3) .. (0,0) .. controls (3.31,0.3) and (6.95,1.4) .. (10.93,3.29)   ;
        
        \draw (248.98,108.48) node [anchor=north west][inner sep=0.75pt]  [rotate=-0.04] [align=left] {$ $};
        \draw (335.79,112.53) node [anchor=north west][inner sep=0.75pt]  [font=\small] [align=left] {$\displaystyle U$};
        \draw (246.07,113.71) node [anchor=north west][inner sep=0.75pt]  [font=\small] [align=left] {$\displaystyle C$};
        \draw (278.8,66.56) node [anchor=north west][inner sep=0.75pt]  [font=\small] [align=left] {$\displaystyle \hat{2}$};
        \draw (205.88,65.89) node [anchor=north west][inner sep=0.75pt]  [font=\small] [align=left] {$\displaystyle \hat{3}$};
        \draw (356.19,65.41) node [anchor=north west][inner sep=0.75pt]  [font=\small] [align=left] {$\displaystyle \hat{1}$};
        \draw (136.99,64.41) node [anchor=north west][inner sep=0.75pt]  [font=\small] [align=left] {$\displaystyle \hat{4}$};
        \draw (242.25,139.88) node [anchor=north west][inner sep=0.75pt]  [font=\small] [align=left] {$\displaystyle x_{c}( T_{\max}) ,v_{c}( T_{\max})$};
        \draw (371.97,151.6) node [anchor=north west][inner sep=0.75pt]  [font=\small,color={rgb, 255:red, 128; green, 128; blue, 128 }  ,opacity=1 ] [align=left] {$\displaystyle L_{f}$};
        \draw (174.46,151.65) node [anchor=north west][inner sep=0.75pt]  [font=\small,color={rgb, 255:red, 128; green, 128; blue, 128 }  ,opacity=1 ] [align=left] {$\displaystyle L_{r}$};
        \draw (386.26,11.19) node [anchor=north west][inner sep=0.75pt]  [color={rgb, 255:red, 208; green, 2; blue, 27 }  ,opacity=1 ] [align=left] {$\displaystyle \overline{S}$};
        \draw (213.86,10.41) node [anchor=north west][inner sep=0.75pt]  [color={rgb, 255:red, 245; green, 166; blue, 35 }  ,opacity=1 ] [align=left] {$\displaystyle \left( i^{*} ,i^{*} +1\right)$};
        \draw (445.8,49.2) node [anchor=north west][inner sep=0.75pt]  [color={rgb, 255:red, 65; green, 117; blue, 5 }  ,opacity=1 ]  {$i^{+}$};
        \draw (37.8,48.8) node [anchor=north west][inner sep=0.75pt]  [color={rgb, 255:red, 65; green, 117; blue, 5 }  ,opacity=1 ]  {$i^{-}$};
        \draw (113,10.57) node [anchor=north west][inner sep=0.75pt]  [color={rgb, 255:red, 65; green, 117; blue, 5 }  ,opacity=1 ] [align=left] {$\displaystyle \overline{S_{C}}$};

    \end{tikzpicture}

    \end{adjustbox}
    \vspace*{-\baselineskip} \vspace*{-2mm}
    \caption{\centering{Cooperative vehicle sets $\bar{S}_C$, $\bar{S}$, and optimal CAV subset $(i^*,i^*+1) \in \bar{S}_C$ selection diagram}}
    \label{fig3:cav_set_selection}
    \vspace*{-\baselineskip}
    \vspace*{-0.2cm}
\end{figure}

Let $S(t)$ be a set of vehicles on the left (fast) lane which are in the vicinity of $C$ at time $t$ and contain all candidate vehicles to cooperate with $C$ in planning its lane-changing maneuver. 
As shown in  Fig. \ref{fig3:cav_set_selection}, we use the parameters $L_r$ and $L_f$ to define this set, where $L_r$ is a given backward distance from the rear end of $C$ and $L_f$ is a forward distance from the front of vehicle $U$. 
Letting $x_i(t)$ denote the longitudinal position of vehicle $i$ along its current lane with respect to a given origin $O$, we define $S(t)$ to consist of fast lane CAVs as follows:
\begin{equation}
    \label{eq8:subset_definition}
        S(t) \coloneqq \left\{i \;\; | \;\; 
        x_C(t)-L_r\leq x_i(t)\leq x_U(t)+L_f  \right\}    
\end{equation}
This set is limited by the communication range between $C$ and other vehicles in its vicinity, but it may otherwise be selected based on any desired criteria. For simplicity, once $N$ members of $S(t)$ are fixed, their indices are ordered and we write $S(t) = \{1,\ldots,N\}$ starting with the CAV furthest ahead of $C$ so that $i+1$ denotes the CAV immediately following $i$ (see Fig. \ref{fig3:cav_set_selection}). 
It is now clear that any pair of vehicles in $S(t)$ selected to cooperate with $C$ a time $t$ is of the form $(i,i+1)$ with $i,i+1 \in S(t)$ 
so that CAV $C$ may merge between any such $i$ and $i+1$.
However, this excludes the possibility of CAV $C$ electing to change lanes ahead of CAV 1 or behind CAV $N$ in this set. Thus, we extend the set to include two more CAVs as follows. 

Let $S^+(t) =  \{ i\; | \; x_i(t) > x_U(t)+L_f   \}$ and $i^+ = \argmin_{i\in S^+(t)} \{ x_i(t) \}$, so that $i^+$ is the CAV immediately ahead of 1 in $S(t)$ if $S^+(t)\neq\emptyset$; else, we view $i^+$ as a ``virtual'' CAV with $x_{i^+}(t) = \infty$. 
Similarly, let $S^-(t) =  \{ i\;\; | \;\; x_i(t) < x_C(t)-L_r   \}$ and $i^- = \arg\max_{i\in S^-(t)} \{ x_i(t) \}$, so that $i^-$ is the CAV immediately behind $N$ in $S(t)$ if $S^-(t) \neq\emptyset$; otherwise, we view $i^-$ as a ``virtual'' CAV with $x_{i^+}(t) = -\infty$. We can now extend $S(t)$ to
\begin{equation}
    \label{eq:subset_definition_C}
        S_C(t) = S(t) \cup \{i^+,i^-\}    
\end{equation}
and we write $S_1(t) = \{0,1,\ldots,N,N+1\}$ by assigning 0 to CAV $i^+$ and $N+1$ to CAV $i^-$. 
An optimal pair, selected as described in the sequel, is, therefore, a subset of this set denoted by $\{i^*,i^*+1\}$ where $i^* \in \{0,1,\ldots,N\}$. The pair $(0,1)$ means that CAV $C$ merges ahead of the first CAV in $S(t)$ and $(N,N+1)$ means that CAV $C$ merges behind the last CAV in $S(t)$.


If multiple maneuvers are to be executed, we index them by $k=1,2,\ldots$ and write $S_C^k(t)$ to represent the associated set corresponding to the specific CAV $C$ that initiates the maneuver. We first limit ourselves to the simpler notation $S_C(t)$.

\subsection{Traffic Flow Speed} 
We define free flow speed as the average speed at which the traffic travels on a lane or a road segment. To design a policy to perform minimally disrupting lane changes, it is necessary to estimate the fast lane free flow speed from the perspective of the ego vehicle $C$. However, due to limitations of observability and sensing range, we can define the flow speed $v_{\text{flow}}$ of the fast lane at time $t$ as a convex combination (with weighting factor $\omega$) of the average speed of the vehicles in the set $S_C(t)$ and the maximum allowable speed of the road. Thus, letting $\Vert S_C(t)\Vert$ be the cardinality of $S_C(t)$, set
\begin{equation}
    v_{S_C}(t) = \dfrac{1}{\Vert S_C(t)\Vert} \sum_{i\in S_C(t)}v_i(t)
    \label{eq:avg_speed}
\end{equation}
and define
\begin{equation}
    v_{\text{flow}}(t) = \omega v_{S_C}(t) + (1-\omega)v_{\max}
    \label{eq:flow_speed}
\end{equation}

\subsection{Vehicle Dynamics}
For every vehicle $i\in S(t)$ its dynamics take the form
\begin{equation}
    \dot{x}_i(t) = v_i(t), \;
    \dot{v}_i(t) = u_i(t)        
    \label{eq:vehicle_dynamics}
\end{equation}
where, in addition to $x_i(t)$, we define $v_i(t)$ and $u_i(t)$ to be vehicle $i$'s velocity and (controllable) acceleration respectively.  Without loss of generality, we define the origin for CAV $i$ involved in a maneuver to be the position $x_C(t_{0})$ of CAV $C$, where $t_{0}$ denotes the time at which the maneuver is triggered. We will use $t_f$ to denote the time when the longitudinal maneuver is complete. The control and speed are constrained as follows:
\begin{equation}
    \begin{matrix}
        u_{i_{\min}}\leq u_i(t)\leq u_{i_{\max}}, \; \; \forall t\in\lbrack t_{0},t_{f}\rbrack
        \\
        v_{i_{\min}}\leq v_i(t)\leq v_{i_{\max}}, \; \; \forall t\in\lbrack t_{0},t_{f}\rbrack
    \end{matrix},
    \label{eq:vehicle_constraints}
\end{equation}
where $v_{i_{\max}}>0$ and $v_{i_{\min}}>0$ denote the maximum and minimum speed allowed, usually determined by the rules of the highway. Similarly, $u_{i_{\max}}>0$ and $u_{i_{\min}}<0$ are $i$'s maximum and minimum acceleration control for a vehicle $i$.

\subsection{Safety Constraints}
Let $\delta_i(v_i(t))$ be the speed-dependent safety distance of CAV $i$, defined as the minimum required distance between $i$ and its immediately preceding vehicle:
\begin{equation}
    \delta_i(v_i(t)) = \varphi  v_i(t)+\varepsilon,
    \label{eq:safety_constraint}
\end{equation}
where $\varphi$ is a constant value that denotes the reaction time (usually defined as $\varphi = 1.8\,s$ \cite{vogel2003comparison}), $\varepsilon$ is a constant, and $\delta_i(v_i(t))$ is specified from the center of $i$ to the center of its preceding vehicle. We can now define all safety constraints that must be satisfied during a lane-changing maneuver of $C$ when cooperating with any two CAVs $(i,i+1)$:
\begin{subequations}
    \begin{align}
        x_{U}(t)-x_{C}(t)&\geq \delta_{C}(v_{C}(t)),\; \; \; \; \, \forall  t\in\lbrack t_0,t_{f}] \label{eq4a:cavC_vehU_safety_constraint}\\ 
        x_{i}(t)-x_{i+1}(t)&\geq \delta_{i+1}(v_{i+1}(t)), \; \; \;\forall t\in\lbrack t_0,t_{f}] \label{eq4b:cavi_cavi1_safety_constraint}\\ 
        x_{C}(t_{f})-x_{i+1}(t_{f})&\geq \delta_{i+1}(v_{i+1}(t_{f})), \label{eq4c:cavC_cavi1_safety_constraint}\\
        x_{i}(t_{f})-x_{C}(t_{f})&\geq \delta_{C}(v_{C}(t_{f})) \label{eq4d:cavi_cavC_safety_constraint}
     \end{align}
    \label{eq:safety_constraint_individual}
    \vspace*{-\baselineskip}
\end{subequations}

\subsection{Traffic Disruption} 
We seek to measure the extent to which a successful lane-changing maneuver may disrupt fast-lane traffic. 
Given any time $t_f>t_0$, let $x_i(t_f)$ be the terminal position of $i$ as determined by some control policy $u_i(t)$, $t\in \lbrack t_0,t_f]$. Thus, for any vehicle $i \in S(t)$, let $D_i(t)$ be a disruption metric defined at a time $t>t_0$ as
\begin{equation}
    D_i(t)=\gamma_x d^i_{x}(t) + \gamma_v d^i_{v}(t),
    \label{eq:disruption-metric-introduction}
\end{equation}
where $d^i_{x}(t)$ and $d^i_{v}(t)$ denote the disruption contribution due to position and speed respectively.
The weights $\gamma_x$ and $\gamma_v$ define a convex combination of the two disruption components and are 
tunable parameters selected to place more emphasis on one or the other form of disruption. 
However, to properly define the dimensionality of the convex combination above, we derive each disruption contribution together with its corresponding normalization factor below.

 {\emph{Position Disruption.}} We define the position disruption $d^i_{x}(t)$ as the square of the disruption caused to $i$ due to an acceleration/deceleration control $u_i(t)$ relative to its undisrupted final position. Thus, we define
\begin{equation}
    d^i_{x}(t) = \left( x_i(t)-\lbrack x_i(t_0)+v_i(t_0)(t-t_0)\rbrack \right )^2
    \label{eq:disruption-due-to-position}
\end{equation}
which specifies the difference between the actual position $x_i(t)$ of $i$ under some control policy at time $t$ and its ideal position obtained by maintaining a constant speed $v_i(t_0)$. This is ``ideal'' because the vehicle's uniform motion is undisrupted, hence also minimizing the energy consumption due to any acceleration/deceleration. 

To normalize $d^i_{x}(t)$ in \eqref{eq:disruption-due-to-position}, we define 
\begin{equation}
    \gamma_x = \frac{\gamma}{(d_{x_{\max}}(t))^2} 
    \label{eq:gamma_x}
\end{equation}
where $\gamma \in [0,1]$ is a tuning parameter and $d_{x_{\max}}(t)$ is the maximum possible position disruption that vehicle $i$ could generate over $[t_0,t_0+t]$, i.e., 
under minimum speed and minimum acceleration (maximum deceleration). Thus, we define two cases: first, the maximum distance traveled under minimum acceleration, and second, the maximum distance traveled under minimum acceleration before the vehicle attains its minimal longitudinal speed. Therefore, $d_{x_{\max}}$ takes the form
\begin{gather}
    d_{x_{\max}}(t)= 
    \begin{cases}
        d_{u_{\min}}, &{\text{if}}\ {u_{\min}(t-t_0)+v_i(t_0)\geq v_{min}}\\
        d_{v_{\min}}, &{\text{otherwise.}}
    \end{cases}
\end{gather}
with 
\begin{equation*}
    d_{u_{\min}} = v_i(t_0)(t-t_0)-\left(v_i(t_0)(t-t_0)+0.5u_{\min}t^2\right)    
\end{equation*}
\begin{multline*}
    d_{v_{\min}}=v_i(t_0)(t-t_0)\\
    -v_{\min}\left((t-t_0)-\dfrac{v_{min}-v_i(t_0)}{u_{\min}}\right)+\frac{v_{i\min}^2-v_i(t_0)^2}{2u_{\min}}
\end{multline*}

{\emph{Speed Disruption.}} We define the speed disruption contribution $d^i_{v}(t)$ as the deviation of the speed $v_i(t)$ of $i$ from the free flow speed $v_{flow}$. Thus, the disruption due to speed is defined as
\begin{equation}
    d^i_{v}(t) = \left(v_i(t)-v_{flow}\right)^2
    \label{eq:disruption-due-to-speed}
\end{equation}
with its corresponding normalization factor defined as
\begin{equation*}
    \gamma_v=\dfrac{1-\gamma}{\max \{(v_{\min}-v_{flow})^2,(v_{\max}-v_{flow})^2\}}.
\end{equation*}
where $\gamma \in [0,1]$ is the same weight parameter in \eqref{eq:gamma_x}.

\emph{Total Disruption.} Let $D_{S}(t)$ be the total disruption for a lane-changing maneuver where each vehicle $i\in S(t)$ involved in the maneuver generates some disruption. Thus, we define the following weighted sum:
\begin{equation}
    D_{S_i}(t)=\sum_{j\in S_i}\zeta_j D_j(t),
    \label{eq:disruption}
\end{equation}
where $S_i\subset S_C(t)$ denotes a subset of $S_C(t)$, defined in \eqref{eq:subset_definition_C}, with the inclusion of CAV $C$. Thus, $S_i$ consists of the vehicle triplet $(C,i,i+1)$ under consideration. However, for notational simplicity, we will refer to the total disruption for each triplet as $D_{S}(t)$. The weights $\zeta_j$, such that $\sum_{j\in S_i}\zeta_j=1$, are introduced to allow for different potential effects that each cooperating CAV will have on the overall total disruption. As an example, if $\zeta_{i+1}$ is large relative to $\zeta_i$ and $\zeta_C$, more weight is placed on the disruption caused to the rear vehicle which allows $C$ to move ahead of it, since that may also affect other vehicles behind $i+1$.

Note that \eqref{eq:disruption} is a quadratic disruption metric that depends only on the total maneuver time length $t_f-t_0$, the terminal positions $x_i(t_f)$, and the terminal speeds $v_i(t_f)$ for every CAV $i$ involved in the maneuver. For a vehicle triplet $\left\{C,\;1,\;2 \right\}$, as depicted in Fig. \ref{Fig1:original_maneuver_diagram}, one can see that $D_{S}(t)$ implicitly penalizes the time that CAV $2$ (which normally decelerates to accommodate $C$) would take to accelerate back to the free flow speed. Further, if the disruption $D_{2}(t)$ for vehicle 2 is large and the vehicle following $2$ is closely behind, then $D_{S}(t)$ implicitly penalizes the deceleration of this vehicle and the time that it would take to accelerate back to the free flow speed. Similarly, if vehicle $C$ merges into the fast lane with a speed lower than the free flow speed, vehicle 2 would have to decelerate to avoid a collision with vehicle $C$ after the lane-changing  maneuver has ended, which can produce a backward slowdown chain reaction. 

Clearly, if there are multiple maneuvers indexed by $k=1,2,\ldots$, we can seek to minimize an aggregate metric
$D_\text{Total}$ by adding individual disruptions $D^k_{S}(t)$.

\subsection{Optimization Problem}
We consider two objectives for the longitudinal maneuver problem: first, we wish to minimize the maneuver time $t_f$ experienced by CAV $C$ and its cooperating vehicles; second, we wish to minimize the energy consumption of each of the three cooperating CAVs, i.e., $C$, $i$ and $i+1$. At the same time, we must satisfy the safety constraints \eqref{eq:safety_constraint} and vehicle constraints \eqref{eq:vehicle_constraints}. Finally, we must ensure that the disruption metric \eqref{eq:disruption} does not exceed a given threshold $D_{th}$. 

The overall optimization problem is outlined next: 

1. Given the initial position for CAV $C$ and vehicle $U$, as well as a maximum allowable time $T_{\max} \geq t_f^*$, we replace the candidate set $S_C(t_f^*)$ by a simpler \emph{fixed} set denoted by $\bar{S}_C$ (to be defined in the sequel) which contains $S_C(t_f^*)$. Then, using the set $\bar{S}_C$,
the free flow speed $v_{flow}$ is estimated as in (\ref{eq:avg_speed}).

2. CAV $C$ determines an optimal terminal time $t_f^*$ and control $\{ u_C^*(t) \}$, $t \in [t_0,t_f^*]$ for the maneuver so as to minimize a given objective function $J_C$ (to be defined in the sequel) subject to the vehicle dynamics (\ref{eq:vehicle_dynamics}), safety constraint 
(\ref{eq4a:cavC_vehU_safety_constraint}), and physical constraints (\ref{eq:vehicle_constraints}). Moreover, its optimal terminal speed $v_C(t_f^*)$ must be close to (or exactly match) the desired speed given by the fast lane speed $v_{flow}$. 

3. The solution $t_f^*$ specifies the terminal position $x_C(t_f^*)$ for CAV $C$. This allows the computation of the trajectory for each possible candidate pair in $\bar{S}_C$. Once each of the optimal trajectories is determined, an optimal pair $(i^*,i^*+1)$ of cooperating CAVs must be selected so as to minimize the disruption metric $D_{S_i}(t_f^*)$ in (\ref{eq:disruption}). Since $D_{S_i}(t_f^*)$ depends on the terminal states $x_i(t_f^*)$, $x_{i+1}(t_f^*)$ in \eqref{eq:disruption}, its minimization depends on the optimal trajectories selected by CAVs $i \in \bar{S}_C$. This requires the determination of optimal controls $\{ u_i^*(t) \}$, $t \in [t_0,t_f^*]$ for all $i \in \bar{S}_C$
minimizing a given objective function $J_i$ (to be defined in the sequel) subject to the vehicle dynamics \eqref{eq:vehicle_dynamics} and constraints
\eqref{eq:vehicle_constraints}, \eqref{eq4b:cavi_cavi1_safety_constraint}, \eqref{eq4c:cavC_cavi1_safety_constraint}, and \eqref{eq4d:cavi_cavC_safety_constraint}. 

4. Finally, we determine an optimal pair $(i^*,i^*+1)$ which minimizes the 
disruption metric $D_{S_i}(t_f^*)$ over all $i \in \bar{S}$. 
This solution must satisfy the requirement $D_{S_i}(t_f^*) \leq D_{th}$. 

A solution to this problem, consisting of $(i^*,i^*+1)$ and $\{ u_C^*(t),u_{i^*}^*,u_{i^*+1}^* \}$, $t \in [t_0,t_f^*]$, may not exist. 
In the next section, we present a detailed solution approach with the following key elements: (i) We specify the objective functions $J_C$ and $J_i$ for any $i \in S_C(t)$. 
(ii) We obtain the optimal cooperating pair $(i^*,i^*+1)$ by evaluating the total disruption that each possible cooperative triplet might produce.  
(iii) We include a time relaxation on $t_f^*$ so that, if a feasible solution does not exist, we seek one for a relaxed value $t'_f > t_f^*$. The relaxation process captures the trade-off between the ``selfish'' goal of $C$ to minimize its maneuver time and the system-wide ``social'' goal of minimizing traffic flow disruptions while ensuring an increase in throughput by adding vehicles to the fast lane.

%% file: sections/long_optimal_control_solution.tex
\begin{figure}[pt]
    \centering
    \vspace*{\baselineskip} \vspace*{-2mm}
    
    \begin{adjustbox}{width=\linewidth, height=7cm,center}    
        \tikzset{every picture/.style={line width=0.75pt}} 
        \begin{tikzpicture}[x=0.65pt,y=0.55pt,yscale=-1,xscale=1]
        
        \draw    (424.28,155.78) -- (418,155.69) ;
        \draw [shift={(416,155.67)}, rotate = 0.79] [color={rgb, 255:red, 0; green, 0; blue, 0 }  ][line width=0.75]    (10.93,-3.29) .. controls (6.95,-1.4) and (3.31,-0.3) .. (0,0) .. controls (3.31,0.3) and (6.95,1.4) .. (10.93,3.29)   ;
        \draw   (424.28,155.78) -- (487.33,155.78) -- (487.33,223) ;
        
        \draw    (276.57,444.42) -- (276.22,472.1) ;
        \draw [shift={(276.19,474.1)}, rotate = 270.74] [color={rgb, 255:red, 0; green, 0; blue, 0 }  ][line width=0.75]    (10.93,-3.29) .. controls (6.95,-1.4) and (3.31,-0.3) .. (0,0) .. controls (3.31,0.3) and (6.95,1.4) .. (10.93,3.29)   ;
        \draw    (275.91,274.07) -- (275.91,306.47) ;
        \draw [shift={(275.91,308.47)}, rotate = 270] [color={rgb, 255:red, 0; green, 0; blue, 0 }  ][line width=0.75]    (10.93,-3.29) .. controls (6.95,-1.4) and (3.31,-0.3) .. (0,0) .. controls (3.31,0.3) and (6.95,1.4) .. (10.93,3.29)   ;
        \draw   (276.31,307.55) -- (370.89,348.59) -- (276.12,389.27) -- (181.54,348.23) -- cycle ;
        \draw    (275.68,389.94) -- (275.88,408.27) ;
        \draw [shift={(275.91,410.27)}, rotate = 269.35] [color={rgb, 255:red, 0; green, 0; blue, 0 }  ][line width=0.75]    (10.93,-3.29) .. controls (6.95,-1.4) and (3.31,-0.3) .. (0,0) .. controls (3.31,0.3) and (6.95,1.4) .. (10.93,3.29)   ;
        \draw   (486.36,311.28) -- (546.67,348.12) -- (486.19,384.74) -- (425.88,347.89) -- cycle ;
        \draw    (370.89,348.59) -- (423.88,349.2) ;
        \draw [shift={(425.88,349.23)}, rotate = 180.66] [color={rgb, 255:red, 0; green, 0; blue, 0 }  ][line width=0.75]    (10.93,-3.29) .. controls (6.95,-1.4) and (3.31,-0.3) .. (0,0) .. controls (3.31,0.3) and (6.95,1.4) .. (10.93,3.29)   ;
        \draw    (486.36,312.61) -- (486.2,276.33) ;
        \draw [shift={(486.19,274.33)}, rotate = 89.75] [color={rgb, 255:red, 0; green, 0; blue, 0 }  ][line width=0.75]    (10.93,-3.29) .. controls (6.95,-1.4) and (3.31,-0.3) .. (0,0) .. controls (3.31,0.3) and (6.95,1.4) .. (10.93,3.29)   ;
        \draw   (546.67,349.46) -- (602,349.46) -- (602,100.68) ;
        \draw    (602,100.68) -- (276.67,100.68) ;
        \draw [shift={(274.67,100.68)}, rotate = 360] [color={rgb, 255:red, 0; green, 0; blue, 0 }  ][line width=0.75]    (10.93,-3.29) .. controls (6.95,-1.4) and (3.31,-0.3) .. (0,0) .. controls (3.31,0.3) and (6.95,1.4) .. (10.93,3.29)   ;
        
        \draw (335.17,198.06) node  [font=\small] [align=left] {\textit{Terminal Time }$\displaystyle t_{f}^{*}$};
        \draw (172.04,199.85) node  [font=\small] [align=left] {\textit{CAV C Terminal Position: }$\displaystyle x_{C}\left( t_{f}^{*}\right)$};
        \draw (175.91,111.35) node  [font=\small] [align=left] {\textit{Set of Candidate Pairs }$\displaystyle S_{i} \in \overline{S}_{C} \ $};
        \draw    (201.07,474.66) -- (350.07,474.66) -- (350.07,503.66) -- (201.07,503.66) -- cycle  ;
        \draw (275.57,489.16) node   [align=left] {\begin{minipage}[lt]{98.49pt}\setlength\topsep{0pt}
        \begin{center}
        \textcolor[rgb]{0.82,0.01,0.11}{\textit{\underline{Execute}}} maneuver
        \end{center}
        
        \end{minipage}};
        \draw (374.26,326.2) node [anchor=north west][inner sep=0.75pt]   [align=left] {\textit{No?}};
        \draw    (141.91,3.33) -- (407.91,3.33) -- (407.91,81.33) -- (141.91,81.33) -- cycle  ;
        \draw (274.91,42.33) node   [align=left] {\begin{minipage}[lt]{178.22pt}\setlength\topsep{0pt}
        \begin{center}
        Find \textcolor[rgb]{0.82,0.01,0.11}{\textit{\underline{all candidate CAV pairs}}} $( i,i+1)$ within range $[ L_{f} ,L_{r}]$ \\from $x_{C}( T_{\max})$ to form $\displaystyle \overline{S}_{C}$
        \end{center}
        
        \end{minipage}};
        \draw    (133.29,125.67) -- (415.29,125.67) -- (415.29,178.67) -- (133.29,178.67) -- cycle  ;
        \draw (136.29,129.67) node [anchor=north west][inner sep=0.75pt]   [align=left] {\begin{minipage}[lt]{188.99pt}\setlength\topsep{0pt}
        \begin{center}
        Find \textcolor[rgb]{0.82,0.01,0.11}{\textit{\underline{energy-optimal time $\displaystyle t_{f}^{*}$ for CAV C}}} \\in slow lane to reach free flow speed $v_{d}$
        \end{center}
        
        \end{minipage}};
        \draw    (413.69,221.55) -- (558.69,221.55) -- (558.69,272.55) -- (413.69,272.55) -- cycle  ;
        \draw (486.19,225.55) node [anchor=north] [inner sep=0.75pt]   [align=left] {\begin{minipage}[lt]{95.87pt}\setlength\topsep{0pt}
        \begin{center}
        \textcolor[rgb]{0.82,0.01,0.11}{\textit{\underline{Relax}}} Terminal Time \\$t'_{f} =t_{f}^{*} \lambda_{t_{f}}$
        \end{center}
        
        \end{minipage}};
        \draw    (145.02,220.05) -- (406.02,220.05) -- (406.02,274.05) -- (145.02,274.05) -- cycle  ;
        \draw (275.52,247.05) node   [align=left] {
        \begin{minipage}[lt]{174.73pt}\setlength\topsep{0pt}
        \begin{center}
            Compute \textcolor[rgb]{0.82,0.01,0.11}{\textit{\underline{Optimal Trajectory}}} for CAV candidate pair$\ ( i,i +1)$ 
        \end{center}
        
        \end{minipage}};
        \draw (335.91,289.68) node  [font=\small] [align=left] {\begin{minipage}[lt]{81.5pt}\setlength\topsep{0pt}
        \begin{center}
        \textit{Disruption }$\displaystyle D_{S_{i}}\left( t_{f}^{*}\right)$
        \end{center}
        
        \end{minipage}};
        \draw    (112.78,410.88) -- (436.78,410.88) -- (436.78,443.88) -- (112.78,443.88) -- cycle  ;
        \draw (274.78,427.38) node   [align=left] {\begin{minipage}[lt]{217.28pt}\setlength\topsep{0pt}
        \begin{center}
        Select \textcolor[rgb]{0.82,0.01,0.11}{\textit{\underline{optimal CAV pair}}} for CAV pair $\displaystyle \left( i^{*} ,i^{*} +1\right)$
        \end{center}
        
        \end{minipage}};
        \draw (350.24,112.68) node  [font=\small] [align=left] {\begin{minipage}[lt]{94.04pt}\setlength\topsep{0pt}
        \begin{center}
        \textit{Free flow speed }$v_{flow}$
        \end{center}
        
        \end{minipage}};
        \draw (287.07,390.83) node [anchor=north west][inner sep=0.75pt]   [align=left] {\textit{Yes?}};
        \draw (275.89,349.23) node  [rotate=-0.11,xslant=0] [align=left] {\begin{minipage}[lt]{84.55pt}\setlength\topsep{0pt}
        \begin{center}
        \textcolor[rgb]{0.82,0.01,0.11}{\underline{\textit{Check Feasibility}}} \\{\small $\displaystyle D_{S_{i}}\left( t_{f}^{*}\right) \leq D_{th}$}
        \end{center}
        
        \end{minipage}};
        \draw (228.17,289.39) node  [font=\small] [align=left] {Trajectories};
        \draw (494.93,290.2) node [anchor=north west][inner sep=0.75pt]   [align=left] {\textit{No?}};
        \draw (558.41,326.23) node [anchor=north west][inner sep=0.75pt]   [align=left] {\textit{Yes?}};
        \draw (486.19,343.41) node  [font=\small,rotate=-0.11,xslant=0] [align=left] {\begin{minipage}[lt]{67.05pt}\setlength\topsep{0pt}
        \begin{center}
        \textit{ Maximum }\\\textit{\# Relaxations?}
        \end{center}
        
        \end{minipage}};
        \draw (514.57,88.68) node  [font=\small] [align=left] {\textit{Increase CAV index} $\displaystyle i$};
        \draw    (274.68,81.33) -- (274.44,123.67) ;
        \draw [shift={(274.43,125.67)}, rotate = 270.33] [color={rgb, 255:red, 0; green, 0; blue, 0 }  ][line width=0.75]    (10.93,-3.29) .. controls (6.95,-1.4) and (3.31,-0.3) .. (0,0) .. controls (3.31,0.3) and (6.95,1.4) .. (10.93,3.29)   ;
        \draw    (274.62,176.67) -- (275.15,218.05) ;
        \draw [shift={(275.18,220.05)}, rotate = 269.26] [color={rgb, 255:red, 0; green, 0; blue, 0 }  ][line width=0.75]    (10.93,-3.29) .. controls (6.95,-1.4) and (3.31,-0.3) .. (0,0) .. controls (3.31,0.3) and (6.95,1.4) .. (10.93,3.29)   ;
        
        \end{tikzpicture}
        
    \end{adjustbox}
        \caption{{Cooperative Maneuver Flow Diagram}}
    \label{Fig2:MinimalDisruptionFlowchart}
    \vspace*{-\baselineskip} \vspace*{-1mm}
    
\end{figure}
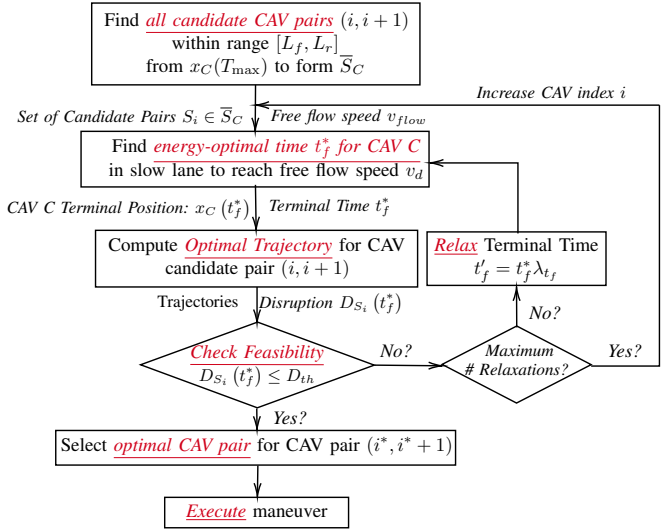

The overall description of the solution process for the optimal maneuver ensuring a minimal disruption that does not exceed a given threshold $D_{th}$ is given in Fig. \ref{Fig2:MinimalDisruptionFlowchart}. 

\subsection{Construction of Cooperative Set $S_C(t)$:} \label{SubSec3_B:optimal_cooperative_pair_selection}

 The set of all candidate fast-lane CAVs with which CAV $C$ can cooperate at any time $t$ was defined in \eqref{eq:subset_definition_C}. We are interested in the set $S_C(t_f^*)$ evaluated at the optimal time for $C$ to complete its longitudinal maneuver. As seen in \eqref{eq8:subset_definition}, this requires not only the value of $t_f^*$ but also the optimal positions of $C$ and all $i$ that satisfy the conditions defining this set. This implies the computation of the associated optimal control problems (specified later in this section). However, this adds an unnecessary level of complexity since we can overestimate the cooperation set by including more potential candidate CAVs without having to compute any optimal control problem until later. In fact, the CAVs that fall into the set defined by the parameters $L_f$ and $L_r$ when the maneuver starts at $t_0$ are not likely to change much, if at all, by the time the maneuver ends. 

Thus, let us start by defining $T_{\max}$ as the maximum acceptable maneuver time so that $t_f^* \leq T_{\max}$. Using $T_{\max}$, we can obtain $S(T_{max})$ in \eqref{eq8:subset_definition} that considers all the vehicles that could interact with CAV $C$ within the time window $[t_0,T_{\max}]$. 
Although the values of $x_i(T_{\max})$ and $x_C(T_{\max})$ in (\ref{eq8:subset_definition}) are still unknown, we can overestimate this set by assuming a constant speed for all times 
$t\in \lbrack t_0 ,\; T_{\max} \rbrack$ , i.e., $x_i(T_{\max}) = x_{i}(t_0)+v_{i}(t_0)T_{\max}$, $x_C(T_{\max}) = x_{C}(t_0)+v_{\min}T_{\max}$. Thus, we define a set 
$\bar{S}$ (omitting the argument $T_{\max}$ for simplicity) as follows:
\begin{equation}
    \label{eq:Sbar}
    \bar{S}\coloneqq  \{ i\;\; | \;\;              
            x_C(T_{\max}) - L_r\leq 
           x_i(T_{\max})\leq 
           x_U(T_{\max})+L_f \}
\end{equation}
where  $x_i(T_{\max})$ is the projected position of $i$ at time $T_{\max}$ under constant speed ($x_{i}(t_0)+v_{i}T_{\max}$), with $T_{\max}$ being the maximum allowable time that a lane-changing maneuver can take. 
We can now augment $\bar{S}$ above by including CAVs $i^+$ and $i^-$ as in (\ref{eq:subset_definition_C}) to obtain the set 
$\bar{S}_C$: 
\begin{equation}
    \label{eq:Sbar_C}
        \bar{S}_C = (\bar{S} \cup \{i^+,i^- \}) 
\end{equation}

\begin{rem} An alternative construction to (\ref{eq:Sbar}) is to first compute $C$'s optimal trajectory and use (\ref{eq8:subset_definition}) with the value of $x_i(t_f^*)$ under maximal acceleration to determine a set $S_{\max}(t_f^*)$ and maximal deceleration 
to determine a set $S_{\min}(t_f^*)$ and then define $S(t_f^*) = S_{\max}(t_f^*) \cup S_{\min}(t_f^*)$. Also, note that by adjusting the parameters $L_f$, $L_r$, $T_{\max}$ the size of this set can be adjusted to include as many candidate CAVs as desired, subject only to the constraint that any $S(t_f^*)$ must include CAVs within the communication range of CAV $C$ to allow full cooperation.
\end{rem}


\subsection{CAV C Optimal Trajectory:} 
Given a CAV $C$ traveling behind an uncontrolled vehicle $U$, a ``start maneuver'' request is sent to surrounding vehicles, and the starting time $t_0$ is defined when $x_U(t)-x_C(t)\leq d_\text{start}$, where $d_\text{start}$ denotes the minimum distance at which CAV $C$  decides to initiate a lane-changing maneuver. It is important to point out that at that instant the values of the optimal maneuver time and the entire optimal trajectory of $C$ can be evaluated, which also enables planning the complete solution to the problem.

We now define the optimal control problem (OCP) for CAV $C$ as follows:
\begin{multline} 
    \label{eq:cav_c_objective_simplified}      
    \min\limits_{t_f,\{u_C(t)\}} w_t\left(t_f-t_0\right)+
    \frac{w_{v}}{2}(v_C(t_f)-v_{flow})^2 \\
    +\int_{t_0}^{t_f} \dfrac{w_u}{2} u_C^2(t) dt
\end{multline}  
\begin{equation*}    
    \begin{matrix}
        \text{s.t. \ \eqref{eq:vehicle_dynamics}, \eqref{eq:vehicle_constraints},}\\
        x_U(t)-x_C(t)\geq \delta_C(v_C(t)),\ \forall t\in [t_0,t_f],\\
        t_0\leq t_f\leq T_{\max}
     \end{matrix}
\end{equation*}
where $w_{\left\{t,~v,~u \right\}}$ are adjustable non-negative weights with appropriate dimensions that place a relative emphasis on each of the three objective function components with respect to each other during the optimization process.
Thus, we can penalize longer maneuver times, deviation from the desired speed, and energy consumption to trade off among these metrics as desired.  
The desired speed value of $v_C(t_f)$ is set to $v_{flow}$ as defined in (\ref{eq:flow_speed}) using 
$v_{\bar{S}_C}$ with $\bar{S}_C$ defined in (\ref{eq:Sbar_C}).
The last two constraints in \eqref{eq:cav_c_objective_simplified} capture (i) the safe distance constraint \eqref{eq4a:cavC_vehU_safety_constraint} between $C$ and $U$ (assuming that the position and speed of $U$ can be sensed or estimated by $C$), and 
(ii) $T_{\max}$ is specified as the maximum tolerable time to perform a lane-changing maneuver. In practice, if the last constraint cannot be met for a given $T_{\max}$, CAV $C$ has the option of either relaxing this value (as detailed in Section \ref{SubSec3_C:Time_Relaxation}) or simply aborting the maneuver. Finally, note that problem (\ref{eq:cav_c_objective_simplified}) is solved given the initial position and speed of CAV $C$.

The solution of this OCP can be analytically obtained, as shown in the Appendix,
through standard Hamiltonian analysis similar to OCPs formulated and solved in \cite{chen2020cooperative}.

It is worth pointing out that depending on the weights $w_i$ and the starting distance $d_\text{start}$ of the maneuver, the form of the corresponding optimal trajectory can be either strictly \emph{accelerating} or first \emph{decelerating} followed by an accelerating component so that \eqref{eq4a:cavC_vehU_safety_constraint} is satisfied regardless of the initial conditions of CAV $C$. Intuitively if there is adequate distance ahead of $C$, it can accelerate at a maximal rate to attain $v_c(t_f) = v_{flow}$; otherwise, it needs to first decelerate to create such an adequate distance ahead of it and then accelerate to minimize the terminal speed cost in (\ref{eq:cav_c_objective_simplified}).

\subsection{Optimal Trajectories for CAV Candidates $\displaystyle (i,i+1)$: }
For any CAV $i$ other than $C$, we define its objective function to be:
\begin{equation*}
        J_{i}(u_i(t)) = 
        \beta(v_i(t^*_f)-v_{flow})^2+
        \int_{t_0}^{t^*_f}\frac{1}{2} {u^2_{i}(t)}dt
    \label{eq:cav_i_objective}
\end{equation*} 
where $t_f^*$ was determined from (\ref{eq:cav_c_objective_simplified}) and
\begin{equation*}
    \beta = \dfrac{\alpha_{v}\max\left\{ {u_{i_{\min}}^2}, {u_{i_{\max}}^2}\right\}}{\left(1-\alpha_{v} \right)}
\end{equation*}
with $\alpha_{v}$ being a constant weight factor with appropriate dimensions that penalizes speed deviation from the desired $v_{flow}$ relative to an energy consumption metric, so that CAV $i$ (where $i=i^*$ or $i=i^*+1$ when the optimal cooperative pair has been found) solves the following two fixed terminal time OCPs:

\subsubsection{CAV $i$}
\begin{equation} 
    \min _{\{u_{i}(t)\}} 
    \beta(v_i(t^*_f)-v_{\text{flow}})^2+
        \int_{t_0}^{t^*_f}\frac{1}{2} {u^2_{i}(t)}dt
    \label{eq:cav_i_problem}
\end{equation}
\vspace*{-4mm}
\begin{gather*}
    \begin{matrix}
        \text{s.t. \ \eqref{eq:vehicle_dynamics}, \eqref{eq:vehicle_constraints}}\\
        x_{i-1}(t)-x_i(t)\geq \delta_i(v_i(t)),\ \forall t\in [t_0,t_f^*],\\
        x_{i-1}(t) = x_{i-1}(t_0)+v_{i-1}(t)(t-t_0)\\
        x_i(t_f^*)-x_C(t_f^*)\geq \delta_C(v_C(t_f^*))
     \end{matrix}
\end{gather*} 
where $x_i(t^*_f)$ is the terminal position for vehicle $i$ under the safety considerations of a potential vehicle ahead of $i$ (labeled $i-1$) for all $ t \in [t_0,t^*_f]$, as well as the safety constraint for the terminal position of CAV $C$ computed by the OCP in \eqref{eq:cav_c_objective_simplified}.

\subsubsection{CAV $i+1$}
\begin{equation} 
    \min _{\{u_{i+1}(t)\}} 
    \beta(v_{i+1}(t^*_f)-v_{flow})^2+
        \int_{t_0}^{t^*_f}\frac{1}{2} {u^2_{i+1}(t)}dt
    \label{eq:cav_i_1_problem}
\end{equation}
\vspace*{-6mm}
\begin{align*}
    \text{s.t. \ \eqref{eq:vehicle_dynamics},}& \text{\ \eqref{eq:vehicle_constraints}}\\
    x_C(t_f^*)-x_{i+1}(t_f^*)&\geq \delta_{i+1}(v_{i+1}(t_f^*))\\
    v_{i+1}(t_f^*)&\geq v_{th}
\end{align*} 
where the terminal position for $i+1$ is only constrained by CAV $C$. Additionally, we remove the safety constraint \eqref{eq4b:cavi_cavi1_safety_constraint} between $i$ and $i+1$ due to Thm. 1 in \cite{chen2020cooperative} where it is shown that in an optimal maneuver CAV $i$ does not accelerate and CAV $i+1$ does not decelerate. 

Lastly, in \eqref{eq:cav_i_1_problem}  we include a terminal constraint on the minimum allowable terminal speed for $i+1$ with the introduction of $v_{th}$, which differs from $v_{flow}$ in that it prevents $i+1$ from reaching a terminal speed which is much lower than $v_{flow}$. In fact, ignoring the $v_{th}$ constraint can lead to a subtle pattern of inefficient maneuvers we observed when implementing our prior controllers in \cite{armijos2022sequential}: when a CAV $i^*+1$ decelerates to a much lower speed than the free traffic flow speed to allow lane changes, it may repeatedly decelerate to allow multiple new maneuvering CAVs to get ahead of it with short maneuver times. This can be very effective in terms of the objective of minimizing maneuver times, at the expense of dramatic traffic disruptions and ultimate high congestion in the fast lane.


\subsection{Selection of Optimal Cooperative Pair $(i^{*},i^{*}+1)$:}
The optimal cooperative pair $(i^*,i^*+1)$ among all $i \in \bar{S}_C$ in (\ref{eq:Sbar}) 
is the one that minimizes the disruption metric in (\ref{eq:disruption}) by selecting the terminal states $\left(x_i(t^*_f),v_i(t^*_f)\right)$ for every pair $i$ and $i+1$ resulting in minimal disruption.

The solution of \eqref{eq:cav_i_problem} and \eqref{eq:cav_i_1_problem} for every pair $(i,i+1)$ respectively provides the optimal terminal positions $x_i(t_f^*)$, $x_{i+1}(t_f^*)$ and terminal speeds $v_i(t_f^*)$, $v_{i+1}(t_f^*)$ that provide the inputs for computing the disruption metric $D_{S_i}(t)$ ,defined in \eqref{eq:disruption}, for the cooperative triplet $S_i \in \bar{S}_C$. We shall denote the minimum disruption as $D^*_{C}(t)$, which corresponds to the triplet $(C,i^*,~i^*+1)$ with minimum disruption. 
Thus, the optimization problem is given as
\begin{equation}
    \begin{aligned}
        (i^*,i^*+1) &= \argmin_{(i,i+1) \in \bar{S}_C} D_{S_i}(t_f^*)\\
        \text{s.t.} &\;\; D_{S_i}(t_f^*) \leq D_{th}
    \end{aligned}
    \label{eq:optimal_cooperative_pair}
\end{equation}
Note, that this is a simple minimization problem defined over the discrete set of candidate vehicles $\bar{S_C}$. Thus, \eqref{eq:optimal_cooperative_pair} is solved by comparing the values of $D_{S_i}(t_f^*)$ obtained over a finite set consisting of vehicle pairs $(i,i+1)$ that satisfy the disruption constraint.

\begin{rem}
If no solution to (\ref{eq:optimal_cooperative_pair}) is found, it is still possible to derive a solution based on the analysis in \cite{chen2020cooperative}, with no consideration of disruption.
Alternatively, we may proceed with the time relaxation process described in Section \ref{SubSec3_C:Time_Relaxation}. 
\end{rem}
\subsection{Maneuver Time Relaxation: } \label{SubSec3_C:Time_Relaxation}
As already mentioned, it is possible that no solution to \eqref{eq:cav_i_problem}, \eqref{eq:cav_i_1_problem}, or \eqref{eq:optimal_cooperative_pair} may be found. The most common reason is due to the fact that the optimal maneuver end time $t_f^*$, determined by CAV $C$ at the first step of the solution approach, is too short to allow $C$ to reach a speed sufficiently close to $v_{\text{flow}}$  and for cooperating CAVs to adjust their positions so as to satisfy the safety constraints in (\ref{eq:safety_constraint}).
In such cases, it is possible to perform a relaxation of $t_f^*$  obtained through \eqref{eq:cav_c_objective_simplified} by trading it off against the energy consumption due to the maneuver extension. Thus, the new terminal time is given as $t'_{f}=t_{f}\lambda_{t_f}$ where $\lambda_{t_f}>1$ is a relaxation factor. Observe that this time modification changes the form of the OCP \eqref{eq:cav_c_objective_simplified}, since the terminal time is now fixed at $t'_{f} > t_f^*$ and the solution will lead to a new terminal position $x_c(t'_f)$ for CAV $C$. The new fixed terminal time OCP formulation is as follows:
\begin{equation} 
    \min _{\{u_{C}(t)\}} 
    \beta(v_{i+1}(t_f')-v_{\text{flow}})^2+
        \int_{t_0}^{t_f'}\frac{1}{2} {u^2_{C}(t)}dt
    \label{eq:cav_c_relaxedTime}
    \vspace*{-\baselineskip}
\end{equation}
\begin{gather*}
    \text{s.t. \ (\ref{eq:vehicle_dynamics}), (\ref{eq:vehicle_constraints}), and}\\
    \begin{aligned} 
        x_{U}(t)-x_{C}(t) &\geq \delta_{C}(v_{C}(t)),\text{ \ \ } \forall  t\in\lbrack t_0,t'_{f}],
    \end{aligned}
\end{gather*} 
This process may continue, as shown in Fig. \ref{Fig2:MinimalDisruptionFlowchart}, until a feasible solution is determined or the constraint $t'_{f} \le T_{\max}$ is violated. Thus, we define a maximum number of iterations allowed for every candidate pair. If the maximum number of iterations is reached, we proceed with the next candidate pair index. 

\begin{rem}\label{rem:time_relaxation}
    Despite time relaxation, problem \eqref{eq:cav_c_relaxedTime} can still be infeasible if $d_\text{start}$ is small or if the constraint \eqref{eq4a:cavC_vehU_safety_constraint}  is active at $t_0$. Therefore, CAV $C$ can abort the maneuver and wait a specified time interval for the next opportunity window. Otherwise, a ``selfish'' maneuver may be performed as in \cite{chen2020cooperative} by computing the minimum feasible terminal time and minimum terminal position for any $i$ and $i+1$ with $i\in \bar{S}_C$.
\end{rem}

%% file: sections/lat_optimal_control_solution.tex
In this paper, the lateral component of the maneuver is no different than the one presented in \cite{chen2020cooperative} for the purely vehicle-centric lane-changing maneuver. In this section, we limit ourselves to an overview of this lateral maneuver component. 

Let $t_{0}^{L}$ be the start time of the lateral phase of the lane-change
maneuver. The most conservative approach is to set $t_{0}^{L}=t_{f}^*$, the
optimal terminal time of the longitudinal phase. However, depending on the \textquotedblleft
aggressiveness\textquotedblright\ of a driver we may select $t_{0}^{L}\leq
t_{f}^*$ as further discussed in this section.

The vehicle dynamics used during the lateral maneuver are expressed as
\begin{figure}[h]
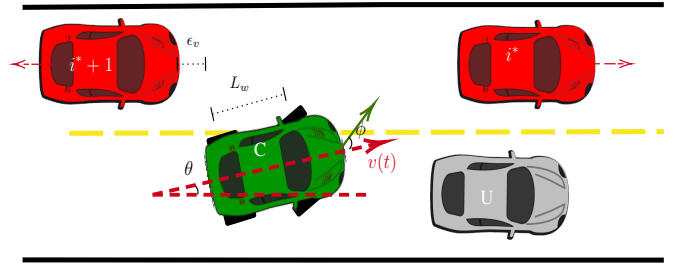

    \vspace*{-\baselineskip}
    \centering
    \begin{adjustbox}{width=\linewidth, height = 4cm,center}
        \tikzset{every picture/.style={line width=0.75pt}} 
        
        \begin{tikzpicture}[x=0.75pt,y=0.75pt,yscale=-1,xscale=1]
        
        \draw [color={rgb, 255:red, 248; green, 231; blue, 28 }  ,draw opacity=1 ][fill={rgb, 255:red, 248; green, 231; blue, 28 }  ,fill opacity=1 ][line width=3]  [dash pattern={on 22.5pt off 7.5pt}]  (57.43,138.11) -- (583.81,136.69) ;
        \draw  [fill={rgb, 255:red, 0; green, 0; blue, 0 }  ,fill opacity=1 ] (179.44,144.38) .. controls (179.2,143.53) and (179.7,142.65) .. (180.55,142.42) -- (203.8,135.97) .. controls (204.65,135.73) and (205.53,136.23) .. (205.76,137.08) -- (207.35,142.8) .. controls (207.59,143.65) and (207.09,144.53) .. (206.24,144.76) -- (182.99,151.22) .. controls (182.14,151.45) and (181.26,150.95) .. (181.03,150.1) -- cycle ;
        \draw  [fill={rgb, 255:red, 0; green, 0; blue, 0 }  ,fill opacity=1 ] (194.44,204.38) .. controls (194.2,203.53) and (194.7,202.65) .. (195.55,202.42) -- (218.8,195.97) .. controls (219.65,195.73) and (220.53,196.23) .. (220.76,197.08) -- (222.35,202.8) .. controls (222.59,203.65) and (222.09,204.53) .. (221.24,204.76) -- (197.99,211.22) .. controls (197.14,211.45) and (196.26,210.95) .. (196.03,210.1) -- cycle ;
        \draw  [fill={rgb, 255:red, 0; green, 0; blue, 0 }  ,fill opacity=1 ] (255.77,198.58) .. controls (255.23,197.88) and (255.36,196.88) .. (256.06,196.34) -- (275.16,181.61) .. controls (275.86,181.07) and (276.86,181.2) .. (277.4,181.9) -- (281.02,186.6) .. controls (281.56,187.3) and (281.43,188.3) .. (280.73,188.84) -- (261.63,203.57) .. controls (260.93,204.11) and (259.93,203.98) .. (259.39,203.28) -- cycle ;
        \draw  [fill={rgb, 255:red, 0; green, 0; blue, 0 }  ,fill opacity=1 ] (241.77,133.58) .. controls (241.23,132.88) and (241.36,131.88) .. (242.06,131.34) -- (261.16,116.61) .. controls (261.86,116.07) and (262.86,116.2) .. (263.4,116.9) -- (267.02,121.6) .. controls (267.56,122.3) and (267.43,123.3) .. (266.73,123.84) -- (247.63,138.57) .. controls (246.93,139.11) and (245.93,138.98) .. (245.39,138.28) -- cycle ;
        \draw (241.09,164.34) node [rotate=-348.35] {\includegraphics[width=91.22pt,height=57.34pt]{greenVehicleTopView.png}};
        \draw [line width=3]    (16,37.87) -- (334.76,35.7) -- (583.81,35.7) ;
        \draw [line width=3]    (16,238.87) -- (297.62,239.03) -- (583.81,239.03) ;
        \draw (438.53,187.25) node  {\includegraphics[width=91.99pt,height=53.37pt]{vehicleTopView.png}};
        \draw (461.97,83.76) node  {\includegraphics[width=129.04pt,height=112.85pt]{redVehicleTopView.png}};
        \draw (92.73,83.5) node  {\includegraphics[width=131.6pt,height=115.95pt]{redVehicleTopView.png}};
        \draw [color={rgb, 255:red, 208; green, 2; blue, 27 }  ,draw opacity=1 ][fill={rgb, 255:red, 208; green, 2; blue, 27 }  ,fill opacity=1 ] [dash pattern={on 3.75pt off 3pt on 7.5pt off 1.5pt}]  (42.82,83.15) -- (9.38,83.15) ;
        \draw [shift={(7.38,83.15)}, rotate = 360] [color={rgb, 255:red, 208; green, 2; blue, 27 }  ,draw opacity=1 ][line width=0.75]    (10.93,-3.29) .. controls (6.95,-1.4) and (3.31,-0.3) .. (0,0) .. controls (3.31,0.3) and (6.95,1.4) .. (10.93,3.29)   ;
        \draw [color={rgb, 255:red, 208; green, 2; blue, 27 }  ,draw opacity=1 ] [dash pattern={on 3.75pt off 3pt on 7.5pt off 1.5pt}]  (522.12,82.77) -- (553.51,82.77) ;
        \draw [shift={(555.51,82.77)}, rotate = 180] [color={rgb, 255:red, 208; green, 2; blue, 27 }  ,draw opacity=1 ][line width=0.75]    (10.93,-3.29) .. controls (6.95,-1.4) and (3.31,-0.3) .. (0,0) .. controls (3.31,0.3) and (6.95,1.4) .. (10.93,3.29)   ;
        \draw [color={rgb, 255:red, 208; green, 2; blue, 27 }  ,draw opacity=1 ][line width=2.25]  [dash pattern={on 6.75pt off 4.5pt}]  (131.2,187.03) -- (325.2,187.03) ;
        \draw [color={rgb, 255:red, 208; green, 2; blue, 27 }  ,draw opacity=1 ][line width=2.25]  [dash pattern={on 6.75pt off 4.5pt}]  (131.2,187.03) -- (332.28,145.84) ;
        \draw [shift={(336.2,145.03)}, rotate = 168.42] [color={rgb, 255:red, 208; green, 2; blue, 27 }  ,draw opacity=1 ][line width=2.25]    (17.49,-5.26) .. controls (11.12,-2.23) and (5.29,-0.48) .. (0,0) .. controls (5.29,0.48) and (11.12,2.23) .. (17.49,5.26)   ;
        \draw [color={rgb, 255:red, 65; green, 117; blue, 5 }  ,draw opacity=1 ][line width=1.5]    (297.2,151.03) -- (325.09,117.18) ;
        \draw [shift={(327,114.87)}, rotate = 129.49] [color={rgb, 255:red, 65; green, 117; blue, 5 }  ,draw opacity=1 ][line width=1.5]    (14.21,-4.28) .. controls (9.04,-1.82) and (4.3,-0.39) .. (0,0) .. controls (4.3,0.39) and (9.04,1.82) .. (14.21,4.28)   ;
        \draw  [draw opacity=0] (304.69,141.21) .. controls (306.37,142.27) and (307.48,144.07) .. (307.49,146.13) .. controls (307.5,147.92) and (306.67,149.52) .. (305.35,150.61) -- (301.25,146.15) -- cycle ; \draw   (304.69,141.21) .. controls (306.37,142.27) and (307.48,144.07) .. (307.49,146.13) .. controls (307.5,147.92) and (306.67,149.52) .. (305.35,150.61) ;  
        \draw  [draw opacity=0] (169,180.19) .. controls (170.43,181.27) and (171.33,182.89) .. (171.34,184.7) .. controls (171.34,185.45) and (171.19,186.16) .. (170.91,186.82) -- (164.79,184.72) -- cycle ; \draw   (169,180.19) .. controls (170.43,181.27) and (171.33,182.89) .. (171.34,184.7) .. controls (171.34,185.45) and (171.19,186.16) .. (170.91,186.82) ;  
        \draw  [dash pattern={on 0.84pt off 2.51pt}]  (153,83.15) -- (178,82.87) ;
        \draw [shift={(178,82.87)}, rotate = 179.35] [color={rgb, 255:red, 0; green, 0; blue, 0 }  ][line width=0.75]    (0,5.59) -- (0,-5.59)   ;
        \draw [shift={(153,83.15)}, rotate = 179.35] [color={rgb, 255:red, 0; green, 0; blue, 0 }  ][line width=0.75]    (0,5.59) -- (0,-5.59)   ;
        \draw  [dash pattern={on 0.84pt off 2.51pt}]  (184,122.87) -- (248,106.87) ;
        \draw [shift={(248,106.87)}, rotate = 165.96] [color={rgb, 255:red, 0; green, 0; blue, 0 }  ][line width=0.75]    (0,5.59) -- (0,-5.59)   ;
        \draw [shift={(184,122.87)}, rotate = 165.96] [color={rgb, 255:red, 0; green, 0; blue, 0 }  ][line width=0.75]    (0,5.59) -- (0,-5.59)   ;
        
        \draw (232.77,150.88) node  [font=\large,color={rgb, 255:red, 255; green, 255; blue, 255 }  ,opacity=1 ] [align=left] {\begin{minipage}[lt]{17.44pt}\setlength\topsep{0pt}
        C
        \end{minipage}};
        \draw (439.26,189.34) node  [font=\large,color={rgb, 255:red, 255; green, 255; blue, 255 }  ,opacity=1 ] [align=left] {\begin{minipage}[lt]{26.64pt}\setlength\topsep{0pt}
        U
        \end{minipage}};
        \draw (83.41,83.15) node  [font=\large,color={rgb, 255:red, 255; green, 255; blue, 255 }  ,opacity=1 ] [align=left] {\begin{minipage}[lt]{38.88pt}\setlength\topsep{0pt}
        $\displaystyle i^{*} +1$
        \end{minipage}};
        \draw (455.91,75.15) node  [font=\large,color={rgb, 255:red, 255; green, 255; blue, 255 }  ,opacity=1 ] [align=left] {\begin{minipage}[lt]{17.8pt}\setlength\topsep{0pt}
        $\displaystyle i^{*}$
        \end{minipage}};
        \draw (309.2,126.9) node [anchor=north west][inner sep=0.75pt]  [font=\large]  {$\phi $};
        \draw (158,158.4) node [anchor=north west][inner sep=0.75pt]  [font=\large]  {$\theta $};
        \draw (320.2,152.4) node [anchor=north west][inner sep=0.75pt]  [font=\large,color={rgb, 255:red, 208; green, 2; blue, 27 }  ,opacity=1 ]  {$v( t)$};
        \draw (160,60.4) node [anchor=north west][inner sep=0.75pt]    {$\epsilon _{v}$};
        \draw (198,91.55) node [anchor=north west][inner sep=0.75pt]    {$L_{w}$};

        \end{tikzpicture}
    \end{adjustbox}
    \caption{Lateral Dynamics Diagram}
    \label{fig:lateral_model}
\end{figure}
\begin{equation}%
    \begin{array}
    [c]{l}%
    \dot{x}(t)=v(t)cos\theta(t)\text{, \ \ }\dot{y}(t)=v(t)sin\theta(t)\\
    \dot{\theta}(t)=v(t)tan\phi(t)/L_{w}\text{, \ \ }\dot{\phi}(t)=\omega(t)
    \end{array}
    \label{lateral_model}%
\end{equation}
where the physical interpretation of all variables above is shown in Fig.
\ref{fig:lateral_model}. In addition, we impose physical constraints as follows:
\begin{equation}
|\phi(t)|\leq\phi_{max}\text{, \ \ }|\theta(t)|\leq\theta_{max}
\label{lateral_constraint}%
\end{equation}
The associated initial conditions are $\phi(t_{0}^{L})=0,$ $\theta(t_{0}%
^{L})=0,$ $y(t_{0}^{L})=0$. The terminal time is defined as $t_{f}^{L}$ and
the associated terminal conditions are
\begin{equation}
\phi(t_{f}^{L})=0\text{, \ \ }\theta(t_{f}^{L})=0\text{, \ \ }y(t_{f}^{L})=l
\label{terminal_lateral}%
\end{equation}
where $l$ is the lane width.

\subsection{Optimal Control Problem Formulation}
Once defined the vehicle's lateral dynamics together with its physical constraints, the optimal control problem for the lateral maneuver is formulated as
\begin{equation}
    \min_{\phi(t),t_{f}^{L}}\text{ }\int_{t_{0}^{L}}^{t_{f}^{L}}\frac{1}{2}%
    w_{\phi}\phi^{2}(t)dt+w_{t_{f}^{L}} t_{f}^{L} \label{lateral_formulation}%
\end{equation}%
\[
    \begin{aligned}  \text{ s.t.  }&\eqref{lateral_model}, \eqref{lateral_constraint}, \eqref{terminal_lateral}
    \end{aligned}
\]
where the objective function combines both the lateral maneuver time and the
associated energy of the controllable vehicle (approximated through the integral of $\phi^{2}(t)$) above. The two terms in
Fig. \ref{fig:lateral_model} need to be properly normalized, therefore, we set
$w_{\phi}=\frac{\rho^{L}}{\phi^{2}_{\max}}$ and $w_{t_{f}^{L}}=\frac
{1-\rho^{L}}{T_{fmax}^{L}}$, where $\rho^{L}\in\lbrack0,1]$ and $T_{fmax}^{L}$
is set based on an empirical value.
 We assume that $v(t)=v$ is constant over the lateral maneuver, which is
reasonable since, as shown in \cite{chen2020cooperative}, the lateral phase time is much smaller compared to
the longitudinal phase. For a complete detailed solution to problem \eqref{lateral_formulation}, we direct the reader to \cite{chen2020cooperative}.

\subsection{Combination of Longitudinal and Lateral Maneuvers}

After addressing the longitudinal and lateral maneuver components separately, we next
	consider how to integrate them into a complete lane change maneuver.
\begin{figure}[h]
	\centering
	\includegraphics[scale=0.36]{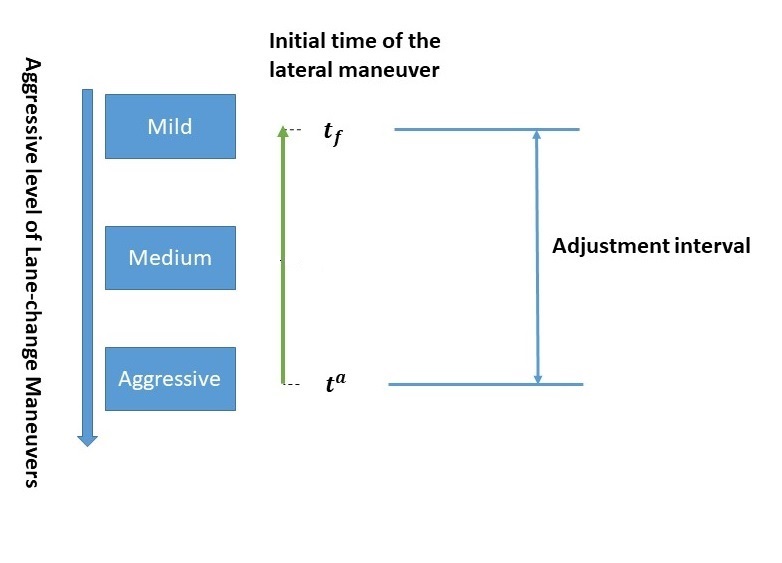} \vspace*{-\baselineskip}
	\vspace*{-\baselineskip}\caption{Maneuver aggressiveness.}%
	\label{aggressive_level}%
\end{figure}
The initial time $t_{0}^{L}$ for the lateral maneuver phase is associated with
a preset driver \textquotedblleft aggressiveness\textquotedblright\ level. As illustrated in Fig.
\ref{aggressive_level}, the most conservative approach is to not
execute the lateral maneuver until the longitudinal phase is complete, i.e.,
set $t_{0}^{L}=t_{f}$. The most aggressive approach is
determined by the earliest time at which CAV $C$ would merge in between $i^*$ and $i^*+1$, that is the time $t^L_0$ at which any adjacent vehicle along the
longitudinal direction can be guaranteed to not collide with 
CAV $C$.

Once the optimal cooperative pair is chosen, let us define the earliest times when CAV $C$ has reached a safe distance form
each of the other three CAVs involved in the longitudinal maneuver in Fig.
\ref{Fig1:original_maneuver_diagram}. Thus, we define $ \tau_{j}$ s.t. $j\in \{i^*,i^*+1,U\} $ as the earliest times at which CAV $C$ has reached a safe terminal position in accordance to \eqref{eq4b:cavi_cavi1_safety_constraint}, \eqref{eq4c:cavC_cavi1_safety_constraint}, and \eqref{eq4a:cavC_vehU_safety_constraint} respectively. Thus, $\tau_{j}$ is defined as follows:
\begin{equation}
    \begin{aligned}
        \tau_{i^*} &   =\min\{t\in\lbrack t_{0},t_{f}]:x_{i^*}(t)-x_{C}(t)\geq \epsilon_{v}\}\\
        \tau_{i^*+1} &   =\min\{t\in\lbrack t_{0},t_{f}]:x_{C}(t)-x_{i^*+1}(t)\geq \epsilon_{v}\}\\
        \tau_{U} &   =\min\{t\in\lbrack t_{0},t_{f}]:x_{U}(t)-x_{C}(t)\geq \epsilon_{v}\} 
    \end{aligned}   
    \label{eq:minimum_time_safety} 
\end{equation}
where $\epsilon_{v}$ denotes a minimum safe distance similar to \eqref{eq:safety_constraint}, typically determined by the length of CAV $C$. We then define the lateral maneuver starting time $t_{0}^{L}=t^a$ as follows:%
\begin{equation}
    t^a  =\max\{\tau_{i^*},\tau_{i^*+1},\tau_{U}\}
    \label{eq:lateral_man_time_start}
\end{equation}

\begin{rem}
   Setting the lateral maneuver starting time $t_{0}^{L}=t^a$ may not always be feasible due to the assumption of constant speeds over the lateral maneuver. However, collisions can still be guaranteed similar to \cref{SubSec3_C:Time_Relaxation}, where the longitudinal maneuver time can be extended to allow larger gaps between $i^*$ and $i^*+1$ and consequently longer adjustment intervals. 
\end{rem}

%% file: sections/sequential_maneuvers.tex
Clearly, we can perform a series of individual maneuvers following Fig. \ref{Fig2:MinimalDisruptionFlowchart} indexed by $k$ that minimize the aggregate metric $D_\text{Total}=\sum^N_{k=1}D^k_{S_i}(t_f^{*k})$. For each maneuver $k$ we can compute a series of system-centric (social) optimal trajectories that start sequentially by defining a CAV $C$ as soon as the maneuver $k-1$ has completed its corresponding lateral phase. Thus, the initial time for maneuver $k$ is upper bounded by the terminal time of maneuver $k-1$ $\left(t^k_0\geq t^{k-1}_f\right)$. Note that CAV $C$ for maneuver $k-1$ can become a CAV candidate for the $k$th maneuver. It is also possible to \emph{parallelize} a number of such maneuvers by allowing them to start simultaneously given a set of target vehicles (CAV $C$). 

%% file: sections/simulation_results.tex
This section provides simulation results illustrating the time and energy-optimal controllers we have derived and comparing their performance against a baseline of non-cooperating (e.g., human-driven) vehicles. Our results are based on using the traffic simulation software package PTV Vissim.

Our simulation setting consists of a straight two-lane highway segment 4000m long and an allowable speed range of $v=\lbrack 10, 35 \rbrack \, m/s$. The incoming traffic 
is spawned with a desired speed of $v_{flow} = 34 \, m/s$. Similarly, the inter-vehicle safe distance \eqref{eq:safety_constraint} is given by $\varepsilon = 1.5\,m$ and headway parameter $\varphi$ drawn from a normal distribution $\mathcal{N}(0.6,\,0.4)\,s$ set to represent tighter bounds due to the assumption of a road composed by 100\% CAVs with communication capabilities. In order to simulate congestion generation, we spawn an uncontrolled vehicle $U$ 
traveling on the right lane (slow lane) with a constant speed $v_U=16\,m/s$ throughout the simulation. The corresponding CAV $C$ is defined as vehicle $U$'s immediately following vehicle. 
For the maneuver start distance we select $d_{\text{start}}$ from $\mathcal{N}(70,10)\,m$ for every CAV $C$ initiating a maneuver. To find the possible CAV candidates on the fast lane, we choose $L_f=50m$ and $L_r=80m$ in setting the candidate set $\bar{S}_C$ in \eqref{eq:Sbar_C}. The control limits specified for every CAV are given by $u_{\min}=-7 \,m/s^2$ and $u_{\max}=3.3 \, m/s^2$. 
The minimum safety distance to perform a lane change was defined as $\epsilon_v=9\,m$ which includes an average vehicle length of $4\,m$. It is assumed that all CAVs in all simulation scenarios share the same parameters and control bounds.
Lastly, we run our simulations on an AMD Ryzen 9 5900x 3.7 GHz. For simplicity, we use CasADi \cite{Andersson2019} as a numerical solver and IPOPT as an interior point optimizer for the computation of the OCP solutions. We employ a numerical solution in order to assess the worst-case time performance for the computation of the trajectories corresponding from solving problems \eqref{eq:cav_c_objective_simplified}, \eqref{eq:cav_i_problem}, \eqref{eq:cav_i_1_problem}, and \eqref{eq:cav_c_relaxedTime} respectively. Clearly, the computational time can be substantially decreased by taking advantage of the analytical solutions derived in \cref{App:HamiltonianAnalysis}.

\subsection{ CAV C Longitudinal Maneuver }

\begin{figure}[hpbt]
    \centering   
    \begin{subfigure}{\linewidth}
    \centering 
      \includegraphics[width=\linewidth, height=7.5cm]{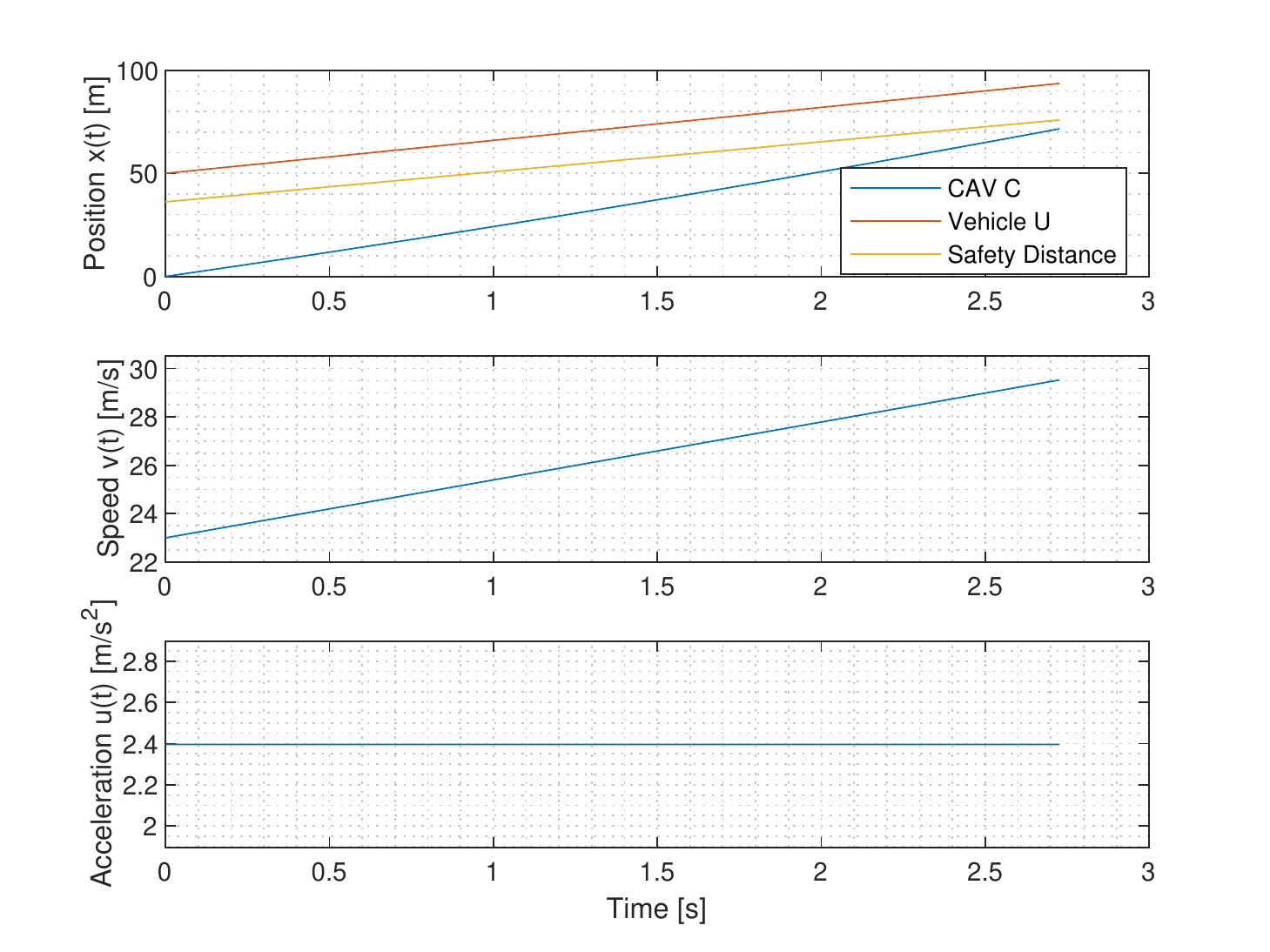}  
      \caption{\centering{Case 1: Constant acceleration sample with no time relaxation}}
      \label{fig6a:acceleration}
    \end{subfigure}   
    
    \begin{subfigure}{\linewidth}
    \centering 
      \includegraphics[width=\linewidth, height=7.5cm]{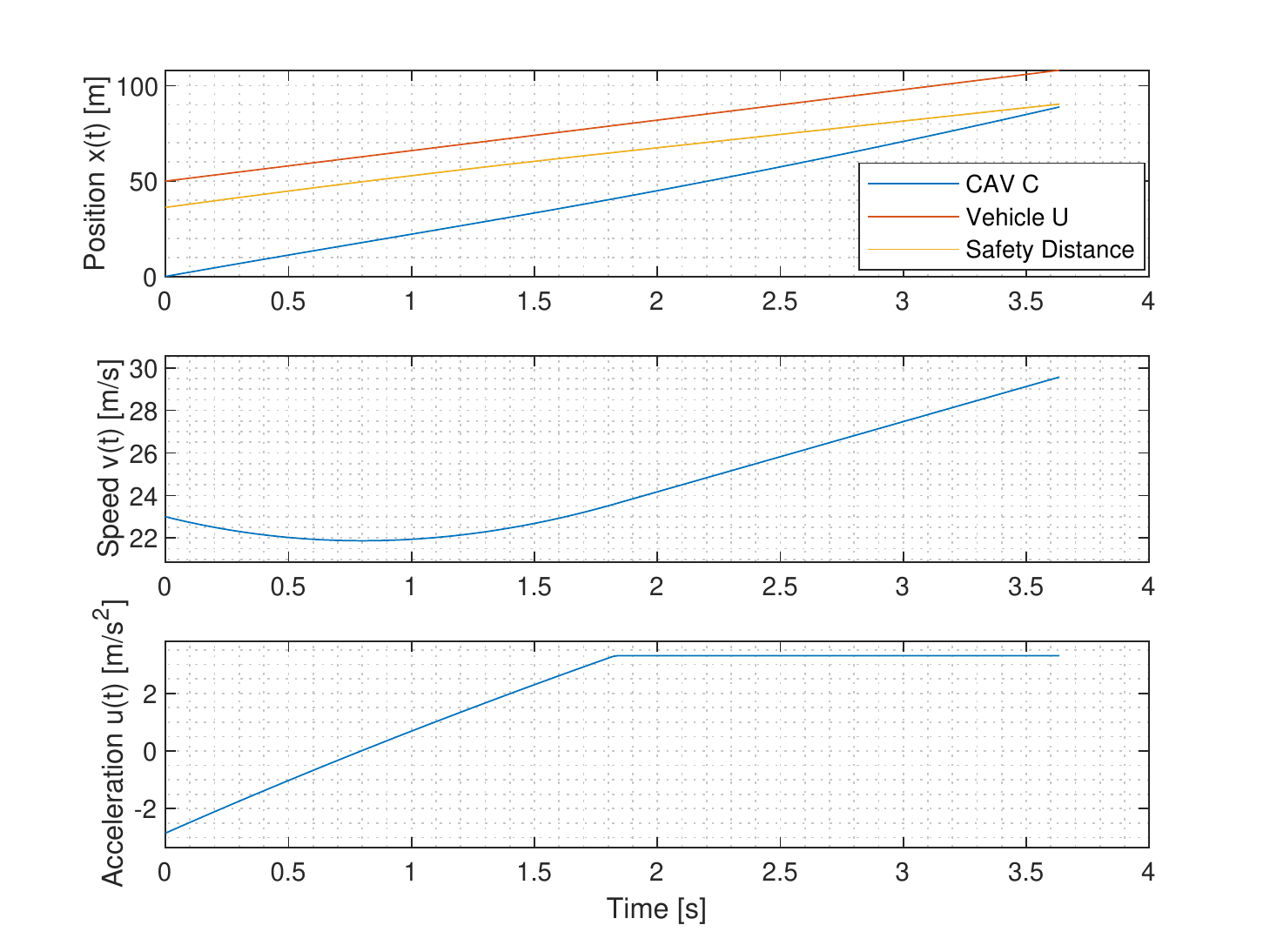}  
      \caption{\centering{Case 2:Mixed acceleration sample under time relaxation}}
      \label{fig6c:mixed_acceleration}
    \end{subfigure}
    \caption{Sample Optimal Trajectory Solutions for CAV $C$}
    \label{fig:CAV_C_AccelerationSamples}  
\end{figure}

We apply the formulation proposed for CAV $C$ in \eqref{eq:cav_c_objective_simplified} for different settings using the desired flow speed $v_{flow}=30\,m/s$. Thus, we provide simulation results for the initial conditions pertaining only to vehicles $U$ and $C$ as described in Table \ref{tab:vehicleCSample} for Cases 1 and 2. For Case 2, we show a sample trajectory generated from the relaxation of the optimal time proposed in \eqref{eq:cav_c_relaxedTime} with the relaxation factor $\lambda_{t_f}=1.1$. Additionally, we define the weighting factors in \eqref{eq:cav_c_objective_simplified} as $w_v=0.50$, $w_t = 0.55$, and $w_u=0.02$.  
 It can be seen for Case 1, when  $d_\text{start}=50\,m$, the resulting maneuver time is $t^*_f=2.7\,s$, and a linear control strategy of constant acceleration to reach $v_{flow}$ is shown in Fig. \ref{fig6a:acceleration}. Conversely, for Case 2, when the resulting relaxed maneuver time is $t^*_f=3.6\,s$ given that CAV $C$ needs to first undergo a deceleration segment to provide enough space to undergo a final acceleration segment allowing CAV $C$ to get close to $v_{flow}$ without violating the safety constraint \eqref{eq4a:cavC_vehU_safety_constraint} as shown in Fig. \ref{fig6c:mixed_acceleration}. Finally, numerically solving the OCP \eqref{eq:cav_c_objective_simplified} with a time discretization of 250 points (this is the hardest problem to solve due to the non-convexity of its objective), 
we obtain results that take an average of $611 \;ms$ to obtain. Similarly, we obtained an average computation time of $85\,ms$ for OCP \eqref{eq:cav_c_relaxedTime} with similar computation times for OCPs \eqref{eq:cav_i_problem} and \eqref{eq:cav_i_1_problem}. 

\begin{table}[pt]
\centering
\vspace*{-\baselineskip} \vspace{-1mm}
\caption{Vehicle C Sample Results}
\label{tab:vehicleCSample}
\resizebox{\linewidth }{!}{%
\begin{tabular}{|c|c|c|c|c|c|c|c|c|}
\hline
    \diagbox[width=10em]{\textbf{{\ul Description}}}{\textbf{{\ul States}}} &
      \textbf{{\ul Relaxed}} &
      \textbf{\begin{tabular}[c]{@{}c@{}}$d_\text{start}$ \\ {[}\textit{m}{]}\end{tabular}} &
      \textbf{\begin{tabular}[c]{@{}c@{}}$x_U(t_0)$\\ {[}\textit{m}{]}\end{tabular}} &
      \textbf{\begin{tabular}[c]{@{}c@{}}$v_U(t_0)$\\ {[}\textit{m/s}{]}\end{tabular}} &
      \textbf{\begin{tabular}[c]{@{}c@{}}$x_C(t_0)$\\ {[}\textit{m}{]}\end{tabular}} &
      \textbf{\begin{tabular}[c]{@{}c@{}}$v_C(t_0)$\\ {[}\textit{m/s}{]}\end{tabular}} &
      \textbf{\begin{tabular}[c]{@{}c@{}}$t_f$\\ {[}\textit{s}{]}\end{tabular}} &
      \textbf{\begin{tabular}[c]{@{}c@{}}$v_C(t_f)$\\ {[}\textit{m/s}{]}\end{tabular}} \\ \hline
    Case 1 & False & 50 & 50 & 16 & 0 & 23 & 2.73  & 29.53 \\ \hline
    Case 2 & True  & 50 & 50 & 16 & 0 & 23 & 3.63  & 29.57 \\ \hline
    \end{tabular}
    }%
\end{table}

\subsection{Sequential Maneuvers } \label{subsec:sequential_maneuvers}

\begin{table*}[pt]
\centering
\caption{Throughput Analysis Summary on a 4000m Highway Segment Under Different Traffic Rates}
\label{tab:finalresultsanalysis}
\resizebox{1\textwidth}{!}{%
\begin{tabular}{|c|c|c|c|c|c|c|c|c|c|c|c|c|} 
\hline
\multirow{2}{*}{\begin{tabular}[c]{@{}c@{}}\textbf{\uline{Traffic Rate}}\\\textbf{\uline{[\textit{veh/hour]}}}\end{tabular}} & \multicolumn{3}{c|}{\textbf{\uline{Throughput [veh/hour] }}}  & \multicolumn{3}{c|}{\textbf{\uline{Maneuver Time [s]}}}   & \multicolumn{3}{c|}{\begin{tabular}[c]{@{}c@{}}\textbf{\uline{Number of Completed Maneuvers}}\\\textbf{\uline{(in 240s)}}\end{tabular}}     & \multicolumn{3}{c|}{\textbf{\uline{Avg. Travel Time [s]}}} \\ 
\cline{2-13}
 & \textbf{\textit{OCMs}} & \textbf{\textit{Baseline}} & \textbf{\textit{Difference [\%]}} & \textbf{\textit{OCMs}} & \textbf{\textit{Baseline}} & \begin{tabular}[c]{@{}c@{}}\textbf{\textit{Difference}}\\\textbf{\textit{[\%]}}\end{tabular} & \textbf{\textit{OCMs}} & \textbf{\textit{Baseline}} & \begin{tabular}[c]{@{}c@{}}\textbf{\textit{Difference}}\\\textbf{\textit{[\%]}}\end{tabular} & \textbf{\textit{OCMs}} & \textbf{\textit{Baseline}} & \begin{tabular}[c]{@{}c@{}}\textbf{\textit{Difference}}\\\textbf{\textit{[\%]}}\end{tabular}  \\ 
\hline
2000    & 1017   & 975    & 4.31             & 6.01    & 29.21    & 79.43   & 26        & 8   & 225.00    & 139.01   & 144.10       & 3.53  \\ 
\hline
3000   & 1251   & 1155   & 8.31              & 6.06  & 44.2  & 86.29  & 36.3            & 5   & 626.00  & 144.52  & 151.04          & 4.32  \\ 
\hline
4000   & 1350  & 1170  & 15.38              & 6.33  & 53.89   & 88.25   & 34.5    
& 4   & 762.50    & 148.37  & 158.77        & 6.55  \\ 
\hline
5000   & 1380  & 1185  & 16.46              & 6.54   & 32.5   & 79.86   & 32            & 6    & 433.33   & 149.02  & 164.28        & 9.29   \\
\hline
\end{tabular}
}
\vspace*{-\baselineskip} 
\end{table*}

We also implemented a series of optimal maneuvers taking into account the system-centric goal of minimizing throughput disruptions. Unlike our previous study in \cite{armijos2022sequential}, the disruption metric now includes a speed disruption component. 

The maximum disruption value used for this study was given as $D_\text{th}=0.15$, with the weight factor $\gamma=0.8$ for position and $1-\gamma=0.2$ for velocity in (\ref{eq:disruption-metric-introduction}). In the optimal control problem (OCP) \eqref{eq:cav_c_objective_simplified} for CAV $C$, the weight factors are given as $w_t = 0.55$, $w_v=0.25$ and $w_u=0.2$ to penalize the maneuver time, terminal velocity costs, and energy consumption respectively. 
 The disruption for CAV $i^*$ is set as 0 since the speed of CAV $i^*$ will never decrease, and the weights $\zeta_C$, $\zeta_{i^*+1}$ for CAV C and CAV $i^*+1$ in (\ref{eq:disruption}) are 0.5 and 0.5, respectively. When calculating the flow speed $v_{flow}$ in (\ref{eq:flow_speed}), we set $\omega = 0.3$ for the weight on the average speed and $1-\omega=0.7$ for the weight on the maximum speed.  When computing the disruption, the maximum disruption allowed was defined to be $D_{th}=0.15$. For simulation purposes, we allowed up to ten relaxations per candidate pair.
The throughput analysis is performed by counting the number of vehicles within a $240\,s$ window that crosses a measurement point at $4000\, m$ from the starting line. To investigate the effectiveness of our controllers over different traffic rates, we performed simulations under traffic rates of 2000, 3000, 4000, and 5000 vehicles/hour, respectively. We collected data to obtain statistics for throughput, maneuver time, number of completed maneuvers, and average travel time for a CAV to pass the specific segment so as to make comparisons to the baseline case of 100\% human-driven vehicles (HDVs) in Vissim.

\begin{figure}[hpbt]
    	\centering
    	\includegraphics[origin=c,scale=0.55]{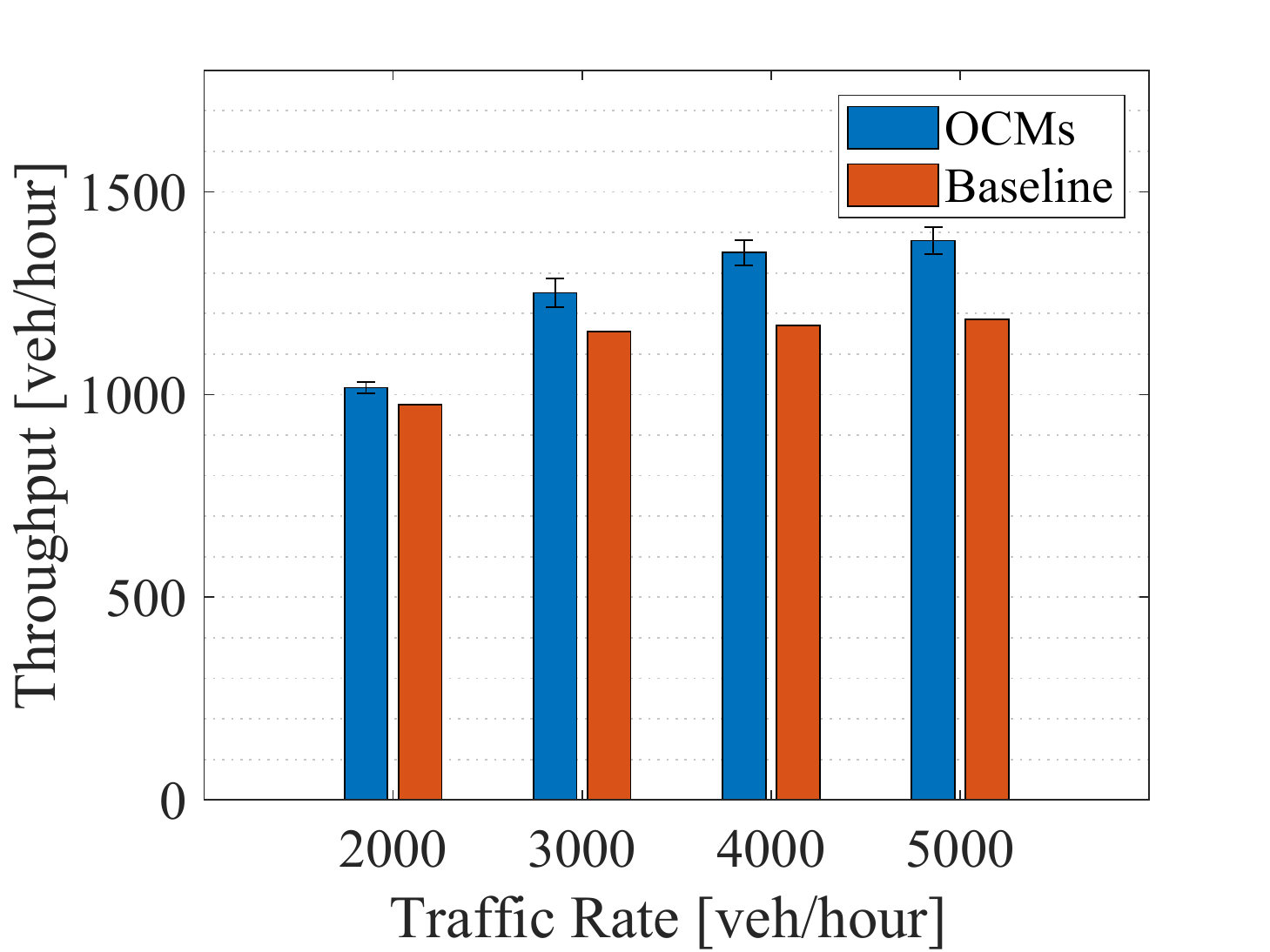}
    	\vspace*{-\baselineskip} \vspace*{\baselineskip}\caption{Throughput Comparison}%
    	\label{fig:throughput}%
        \vspace*{-\baselineskip*2} 
    \end{figure}

\begin{figure}[hpbt]
    	\centering
    	\includegraphics[origin=c,scale=0.55]{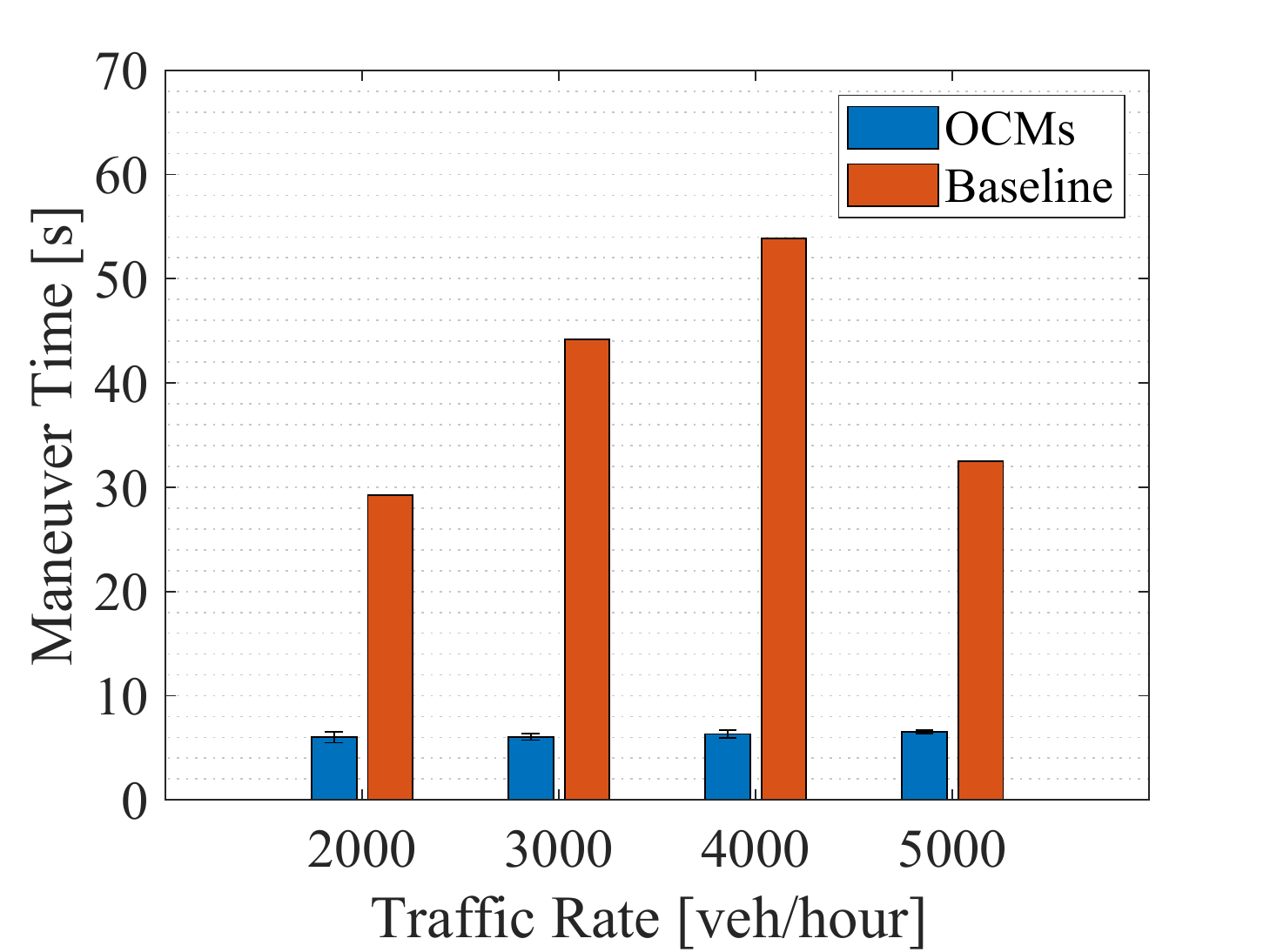}
    	\vspace*{-\baselineskip} \vspace*{\baselineskip}\caption{Maneuver Time Comparison}%
    	\label{fig:tf}%
    \end{figure}

\begin{figure}[hpbt]
    \vspace*{-\baselineskip}
    	\centering
    	\includegraphics[origin=c,scale=0.55]{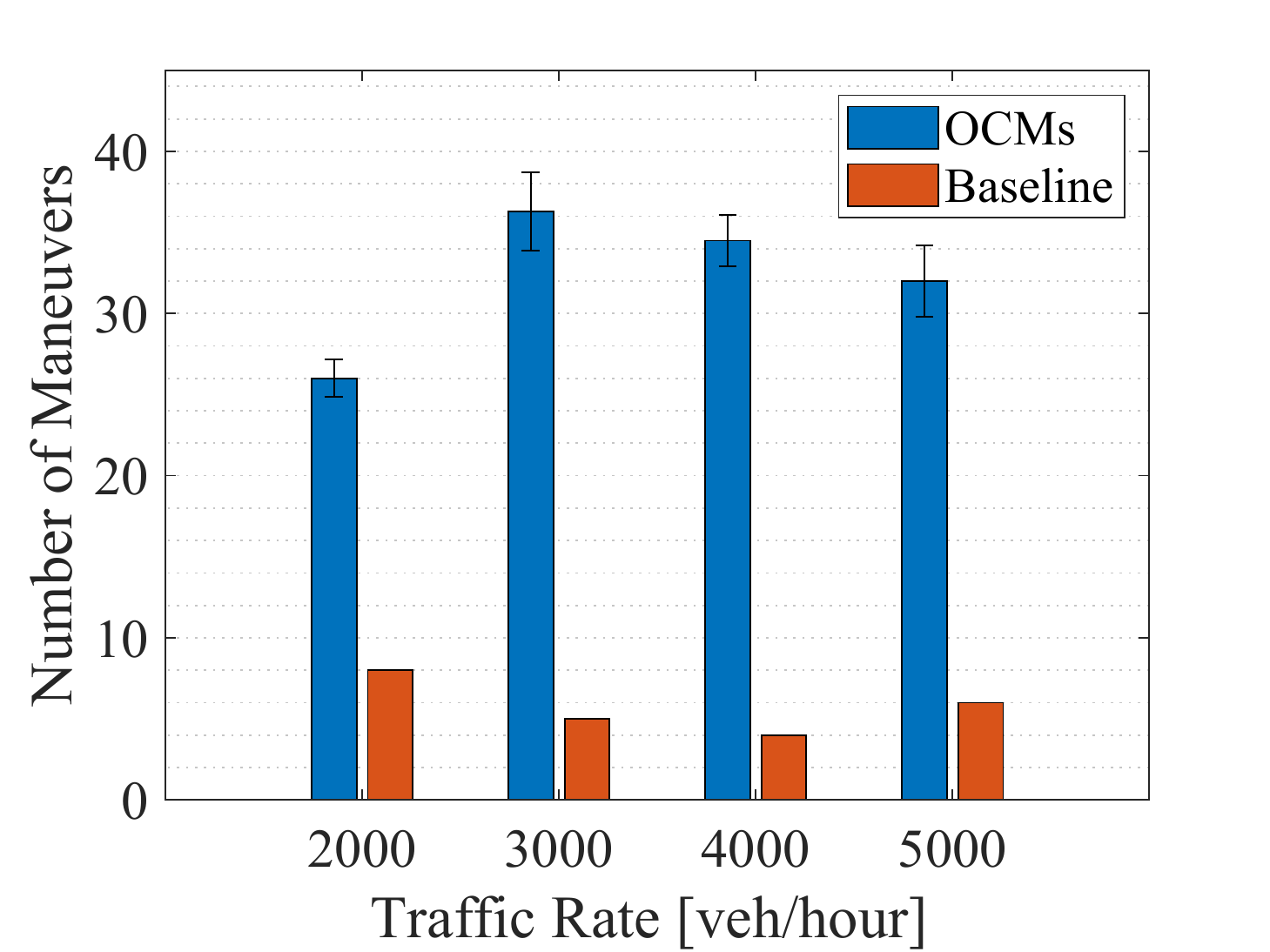}
     \caption{Number of Completed Maneuvers Comparison}%
    	\label{fig:maneuverNumber}%
    \end{figure}

The performance results for optimally controlled CAVs and non-cooperating vehicles under different traffic rates are summarized in Table \ref{tab:finalresultsanalysis}. We use ``OCMs'' and ``Baseline'' to represent cooperative Optimal Control Maneuvers and non-cooperative human-driven cases, respectively. Table \ref{tab:finalresultsanalysis} shows that the throughput improvement of our OCMs strategies increases from 4.31\% to 16.46\% compared to the baseline as the traffic rates increase from 2000 to 5000. Meanwhile, the maneuver time of CAV $C$ is around 7\text{s} in the OCMs, while it takes a vehicle without any cooperation benefit more than $30\text{s}$ to complete a lane change maneuver in the baseline. The average maneuver time for CAVs decreases by around 80\%, which in turn reduces vehicle energy consumption when a lane change is performed. Additionally, the number of CAVs completing the maneuver in OCMs increases by at least 225\% and up to 762.5\%, and the average travel time for a CAV to pass the segment is reduced by 3.53\% to 9.29\% when traffic rates increase from 2000 to 5000 respectively. Moreover, we provide bar charts for the throughput, maneuver time, and the number of CAV $C$ maneuvers comparing the different performance metrics under several traffic rates as shown in Figs. \ref{fig:throughput}, \ref{fig:tf}, and \ref{fig:maneuverNumber} respectively. These include the mean value and standard deviation under different traffic rates for both OCMs and baseline cases.  The differences for each pair in these charts are statistically significant, indicating that the optimally controlled CAVs experience significantly improved baseline performance, while also resulting in an improved OCMs throughput as well.


\subsection{Sensitivity Analysis}

\begin{figure}[pbt]
\vspace*{-4\baselineskip} 
    \centering
    \includegraphics[origin=c, width=1.05\linewidth]{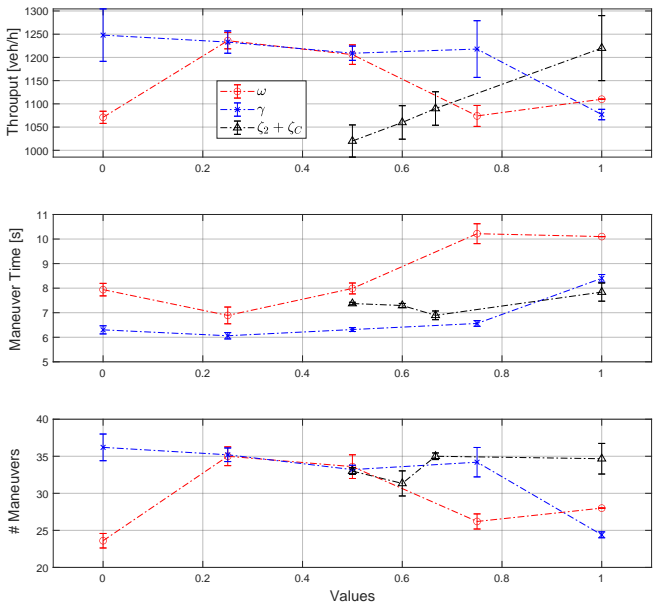}
    \vspace*{-6\baselineskip} 
    \caption{Sensitivity Analysis}
    \vspace*{-5mm}
    \label{fig:sensitivity_analysis}
    
\end{figure}

In this section, we wish to quantify the effect that each of the several parameters used in our analysis can have on the performance of an individual maneuver, as well as their impact on the overall traffic network performance. Thus, we have performed a sensitivity analysis on the maneuver times and the number of maneuvers as the individual (vehicle-centric) performance metrics. Similarly, we analyzed the average throughput on each simulation run as a metric for traffic network (system-centric) performance. Specifically, using a  similar simulation setup as in \cref{subsec:sequential_maneuvers} with a total simulation length of 240 seconds, we performed three sets of experiments by modifying parameters $\omega$ defined in \eqref{eq:avg_speed}, $\gamma$ defined in \eqref{eq:disruption-metric-introduction}, and $\zeta_i$ defined in \eqref{eq:disruption}. A summary containing the parameters used for the sensitivity analysis is shown in Table \ref{tab:sensitivity_parameters_summary}. Similarly, the averaged results over each corresponding set of parameters are shown in Fig. \ref{fig:sensitivity_analysis}.

\begin{table}[hpbt]
    \centering
    \caption{Parameter Summary}
    \label{tab:sensitivity_parameters_summary}
    \resizebox{\columnwidth}{!}{%
    \begin{tabular}{|c|c|c|c|c|c|c|c|c|}
    \hline
    \textbf{\begin{tabular}[c]{@{}c@{}}Traffic Density \\ $\lbrack veh/h \rbrack$\end{tabular}} & \textbf{\begin{tabular}[c]{@{}c@{}}Number \\ Runs\end{tabular}} & {\ul \textbf{$D_{th}$}} & {\ul \textbf{$\omega$}} & {\ul \textbf{$\gamma$}} & {\ul \textbf{$\zeta_1$}} & {\ul \textbf{$\zeta_2$}} & {\ul \textbf{$\zeta_C$}} & {\ul \textbf{$\zeta_2+\zeta_C$}} \\ \hline
    3000 & 5 & 0.15 & {\color[HTML]{FE0000} 0} & 0.8 & 0 & 0.5 & 0.5 & 1 \\ \hline
    3000 & 5 & 0.15 & {\color[HTML]{FE0000} 0.25} & 0.8 & 0 & 0.5 & 0.5 & 1 \\ \hline
    3000 & 5 & 0.15 & {\color[HTML]{FE0000} 0.5} & 0.8 & 0 & 0.5 & 0.5 & 1 \\ \hline
    3000 & 5 & 0.15 & {\color[HTML]{FE0000} 0.75} & 0.8 & 0 & 0.5 & 0.5 & 1 \\ \hline
    3000 & 5 & 0.15 & {\color[HTML]{FE0000} 1} & 0.8 & 0 & 0.5 & 0.5 & 1 \\ \hline
    3000 & 5 & 0.15 & 0.3 & {\color[HTML]{FE0000} 0} & 0 & 0.5 & 0.5 & 1 \\ \hline
    3000 & 5 & 0.15 & 0.3 & {\color[HTML]{FE0000} 0.25} & 0 & 0.5 & 0.5 & 1 \\ \hline
    3000 & 5 & 0.15 & 0.3 & {\color[HTML]{FE0000} 0.5} & 0 & 0.5 & 0.5 & 1 \\ \hline
    3000 & 5 & 0.15 & 0.3 & {\color[HTML]{FE0000} 0.75} & 0 & 0.5 & 0.5 & 1 \\ \hline
    3000 & 5 & 0.15 & 0.3 & {\color[HTML]{FE0000} 1} & 0 & 0.5 & 0.5 & 1 \\ \hline
    3000 & 3 & 0.15 & 0.5 & 0.8 & {\color[HTML]{FE0000} 0.5} & {\color[HTML]{FE0000} 0.5} & {\color[HTML]{FE0000} 0} & {\color[HTML]{329A9D} 0.5} \\ \hline
    3000 & 3 & 0.15 & 0.5 & 0.8 & {\color[HTML]{FE0000} 0.4} & {\color[HTML]{FE0000} 0.2} & {\color[HTML]{FE0000} 0.4} & {\color[HTML]{329A9D} 0.6} \\ \hline
    3000 & 3 & 0.15 & 0.5 & 0.8 & {\color[HTML]{FE0000} 0.33} & {\color[HTML]{FE0000} 0.33} & {\color[HTML]{FE0000} 0.34} & {\color[HTML]{329A9D} 0.67} \\ \hline
    3000 & 3 & 0.15 & 0.5 & 0.8 & {\color[HTML]{FE0000} 0} & {\color[HTML]{FE0000} 0.5} & {\color[HTML]{FE0000} 0.5} & {\color[HTML]{329A9D} 1} \\ \hline
    \end{tabular}%
    }
\end{table}

It can be seen from Fig. \ref{fig:sensitivity_analysis} that for the position disruption weight $\gamma$, the overall number of performed maneuvers did not change much on average. However, it can be seen that whenever $\gamma\geq 0.75$ the throughput starts seeing a dramatic decrease in magnitude. On the contrary, with increasing $\gamma$, a monotonic increase in the average maneuver time can be observed, with a particular emphasis when the speed disruption contribution is almost null. This shows the overall need for the speed disruption contribution. 

On a similar note, it can be seen that the weight $\omega$ used in the flow speed estimate plays a significant role in the average number of maneuvers performed. Similarly, it  can be seen that the average throughput behaves proportionally with the average number of maneuvers influenced by the changes in $\omega$. The optimal behavior can be seen to occur whenever $\omega\approx 0.25$, meaning that a larger weight should be given to $v_{\max}$. The aforementioned behavior can be explained given that the average traffic speed is only given by a smaller subset $\bar{S}_C$ of the high-speed lane which might have been disrupted by a previous maneuver, but the overall traffic flow speed might be much higher. Nevertheless, whenever only using $v_{flow}=v_{\max}$ (or $\omega=0$), it can be seen that the overall maneuver length is increased significantly with a proportional decrease in throughput. 

Lastly, the overall effect of $\zeta_i$ in (\ref{eq:disruption}) is analyzed by varying the split contributions of every CAV $C$, $i^*$, and $i^*+1$ involved in each maneuver. It can be seen that the effect on varying $\zeta_i$ does not impact the average number of maneuvers for each run. However, it can be observed from the throughput analysis that the behavior is linear with the highest throughput obtained when $\zeta_1=0$. Such high throughput, however, comes at an expense of slightly higher maneuver times. This phenomenon is consistent with Theorem 1 in \cite{chen2020cooperative} where CAV $i^*$ can only accelerate or maintain its speed, thus making its possible disruption contribution negligible. 

%% file: sections/Conclusions.tex
We have developed a decentralized optimal control framework for multiple cooperating CAVs 
that combines the ``vehicle-centric'' objective of
minimizing the maneuver time and energy consumed by all cooperating CAVs
and the ``system-centric'' objective of minimizing throughput disruption in the fast lane. Our analysis includes the selection of an optimal cooperation pair of CAVs within a neighboring candidate set that minimizes a disruption metric for the fast lane traffic flow to ensure it never exceeds a given threshold. 
Simulation results show the effectiveness of the proposed controllers with improvements of up 
to 16\% and 90\% in average throughput and maneuver time respectively when compared to maneuvers with no vehicle cooperation.
Ongoing work aims to perform  multiple maneuvers simultaneously while still minimizing traffic disruption. 
An important next step is to extend our analysis to a mixed-traffic setting with both CAVs and human-driven vehicles (HDVs).

%% file: sections/appendix.tex
\subsection{CAV C Trajectory Hamiltonian Analysis}\label{App:HamiltonianAnalysis}

The solution of the OCP (\ref{eq:cav_c_objective_simplified}) for CAV $C$ can be analytically obtained by standard Hamiltonian analysis. To simplify the expression of the objective, we set the parameters 
\begin{align*}
    &\alpha_t=\dfrac{w_t}{w_u},
    \; \alpha_v=\dfrac{w_v}{w_u}
\end{align*}
where $w_{\{t,v,u\}}$ are adjustable non-negative weights in (\ref{eq:cav_c_objective_simplified}). The complete problem is restated below:
\begin{multline*}
    \label{appeq:cav_c_objective_simplified}   
    \min\limits_{t_f,u_C(t)} \alpha_t\left(t_f-t_0\right)+
    \frac{\alpha_{v}}{2}(v_C(t_f)-v_{flow})^2+
    \int_{t_0}^{t_f}\frac{1}{2}u_C^2(t) dt
\end{multline*}
\begin{equation}
    \begin{matrix}
        \dot{x}_C(t) = v_C(t), \;
    \dot{v}_C(t) = u_C(t), \; \; 
        \label{appeq:vehicle_dynamics} \\
    u_{C_{\min}}\leq u_C(t)\leq u_{C_{\max}}, \; \; \forall t\in\lbrack t_{0},t_{f}\rbrack
        \\
        v_{C_{\min}}\leq v_C(t)\leq v_{C_{\max}}, \; \; \forall t\in\lbrack t_{0},t_{f}\rbrack,\\
        x_U(t)-x_C(t)\geq \delta_C(v_C(t)),\ \forall t\in [t_0,t_f],\\
        0\leq t_f\leq T_{\max}
     \end{matrix}
\end{equation}
Let $\mathbf{x_C(t)}:=(x_C(t),v_C(t))^T$ be the state vector and $\mathbf{\lambda_C(t)}=(\lambda_C^x(t),\lambda_C^v(t))^T$ be the costate vector. The Hamiltonian for (\ref{appeq:vehicle_dynamics}) with the state constraint, control constraint, and safety constraint adjoined is
\begin{multline}
    \label{eq:hamiltonian}
    H(\mathbf{x_C},\mathbf{\lambda_C},u_C)=
    \frac{1}{2}u_C^2(t)+\alpha_t+\lambda_C^x(t)v_C(t)+\\
    \lambda_C^v(t)u_C(t)+\mu_1(u_{C,\min}-u_C(t))+\\
    \mu_2(u_C(t)-u_{C,\max})+\mu_3(v_{C,\min}-v_C(t))+\\
    \mu_4(v_C(t)-v_{C,\max})+\mu_5(x_C(t)+\varphi v_C(t)+\varepsilon-x_U(t)).
\end{multline}
The Lagrange multipliers $\mu_1,\mu_2,\mu_3,\mu_4,\mu_5$ are positive when their corresponding constraints are active and become 0 when the constraints are inactive. Note that the problem has an unspecified terminal time $t_f$, and the terminal condition for $t_f$ is contained in the objective. From \cite{bryson2018applied}, the terminal cost is given as $\Phi:=\frac{\alpha_v}{2}[v_C(t_f)-v_{flow}]^2$ which is not an explicit function of time, and the transversality condition for (\ref{eq:hamiltonian}) is 
\begin{align}
\label{eq:transversality}
(\Phi_t+H)|_{t=t_f}=H(\mathbf{x_C}(t),\mathbf{\lambda_C}(t),u_C(t))|_{t=t_f}=0,
\end{align}
with the costate boundary conditions
\begin{align*}
       &\lambda_C^x(t_f)=(\dfrac{\partial\Phi}{\partial x_C})|_{t=t_f}=0,\\
       &\lambda_C^v(t_f)=(\dfrac{\partial\Phi}{\partial v_C})|_{t=t_f}=\alpha_v[v_C(t_f)-v_{flow}].
\end{align*}
The Euler-Lagrange equations are given as 
\begin{align}
\label{eq:eular_lagrange}
\nonumber\dot{\lambda}_C^x&=-\dfrac{\partial H}{\partial x_C}=-\mu_5,\\
\dot{\lambda}_C^v&=-\dfrac{\partial H}{\partial v_C}=-\lambda_C^x+\mu_3-\mu_4-\varphi \mu_5,
\end{align}
and the necessary condition for optimality is 
\begin{equation}
\label{appeq:optimality_condition}
\dfrac{\partial H}{\partial u_C}=u_C(t)+\lambda_C^v(t)-\mu_1+\mu_2=0.
\end{equation}

\subsubsection{Control, state, safety constraints inactive}

In this case, the constraints in (\ref{appeq:vehicle_dynamics}) are inactive for all $t\in [t_0,t_f]$, and we have $\mu_1=\mu_2=\mu_3=\mu_4=\mu_5=0$. Applying the Euler-Lagrange equations in (\ref{eq:eular_lagrange}), we get $\dot{\lambda}_C^x=-\mu_5=0$ and  $\dot{\lambda}_C^v=-\lambda_C^x(t),$ which imply that $\lambda_C^x=a$ and $\lambda_C^v=-(at+b)$, respectively. The parameters $a,b$ here are integration constants. From (\ref{appeq:optimality_condition}), we have
\begin{equation}
\label{appeq:costate_v}
u_C(t)+\lambda_C^v(t)=0,
\end{equation}
and $u_C(t)=-\lambda_C^v(t)=at+b$. Moreover, considering the boundary condition of the costate vector at time $t_f$, we have 
\begin{equation}
\label{appeq:costate_x}
\lambda_C^x(t_f)=\dfrac{\partial \Phi}{\partial x}|_{t=t_f}=0,
\end{equation}
which indicates that $\lambda_C^x(t)=a=0$ for all $t\in [t_0,t_f]$, and we get $\lambda_C^v(t)=-b$ and $u_C(t)=b$ for all $t\in[t_0,t_f]$. Furthermore, we have $\alpha_v [v_{flow}-v_C(t_f)]=b$ according to the costate boundary condition.

From (\ref{eq:transversality}), the transversality condition gives the following relationship 
\begin{equation}
    \label{appeq:transversality_relationship}
    \dfrac{1}{2}u_C^2(t_f)+\alpha_t+\lambda_C^x(t_f)v_C(t_f)+\lambda_C^v(t_f)u_C(t_f)=0,
\end{equation}
and recalling that $u_C(t)=b$, it follows that $b=\pm\sqrt{2\alpha_t}$ and $u_C(t)=\pm\sqrt{2\alpha_t}$. Consequently, we obtain the following optimal solution for $t\in[t_0,t_f]$:
\begin{align}
u_C^*(t)&=b=\pm\sqrt{2\alpha_t},\\
\label{appeq:unconstrained_opt_v}
v_C^*(t)&=v_0\pm\sqrt{2\alpha_t}(t-t_0),\\
x_C^*(t)&=x_C(t_0)+  v_0(t-t_0)\pm\frac{1}{2}\sqrt{2\alpha_t}(t-t_0)^2.
\end{align}
Furthermore, the costate condition and (\ref{appeq:unconstrained_opt_v}) provide the terminal time $t_f$ as
\begin{equation}
    \label{appeq:opt_tf}
t_f^*=t_0+\dfrac{\alpha_v (v_{flow}-v_0)\mp\sqrt{2\alpha_t}}{\pm\alpha_v\sqrt{2\alpha_t}}.
\end{equation}
Observe that in this case, the optimal control input is a time-invariant acceleration. The sign of the optimal control $u_C^*(t)$ depends on the initial speed $v_C(t_0)$.

\subsubsection{Some constraints active}

Define $\bar{x}_C(t_f)$ as the terminal position of CAV $C$ if $u_C(t)=0$ for all $t\in[t_0,t_f]$, and $x_C(t_f)$ as the actual terminal position of CAV $C$. Then, we can specify several cases depending on the relationship between $\bar{x}_C(t_f)$ and $x_C(t_f)$. 
Following the same analysis as in \cite{chen2020cooperative}, there only exist three feasible cases under which the safety constraints are satisfied: \\
\\
\textbf{Case 1:} $\bar{x}_C(t_f)\leq x_C(t_f)\leq x_U(t_f)-\delta_C$

If $\bar{x}_C(t_f)\leq x_C(t_f)$, then $u^*_C(t)\geq 0$ and $u_C(t)=-\sqrt{2\alpha_t}$ is infeasible. For $u_C(t)=\sqrt{2\alpha_t}$, we consider whether the velocity, acceleration, and safety constraints are active or not. 

Firstly, we consider the control constraint. Since $u_C(t)=\sqrt{2\alpha_t}$ is a constant acceleration which only depends on $\alpha_t$, we can easily compare it with $u_{C,\max}$ and $u_{C,\min}$. If $\sqrt{2\alpha_t}> u_{C,\max}$, then $u^*_C(t)=u_{C,\max}$; if $\sqrt{2\alpha_t}< u_{C,\max}$, then the acceleration constraint will never be activated in this case. 

Secondly, we consider the velocity constraint. Since CAV $C$ accelerates to $v_{flow}$ with constant acceleration $u_C(t)=\sqrt{2\alpha_t}$, we only need to compare $v_C(t_f)$ to $v_{flow}$. Since the desired velocity $v_{flow}$ has to satisfy $v_{flow}\leq v_{C,\max}$ and we have $u_C(t)\geq 0$, it follows that the velocity constraint will never be activated in this case. 

Finally, we check the safety constraint. Suppose at time $t_1\in[t_0,t_f)$ the safety constraint is active under $u_C(t)=\sqrt{2\alpha_t}$. Then, $x_C(t_1)=x_U(t_1)-[\varphi v_C(t_1)+\varepsilon]$. At time $t_1^-$, the position of CAV $C$ satisfies $x^*_C(t_1^-)<x_U(t_1^-)-[\varphi v_C(t_1^-)+\varepsilon]$, and we must have $v_C(t_1^-)>v_U$ to activate the safety constraint. With the continuity of $x_C(t),v_C(t)$ and $u_C(t)\geq 0$, the position of CAV $C$ at time $t_1^+$ should satisfy $x_C^*(t_1^+)>x_U(t_1^+)-[\varphi v_C(t_1^+)+\varepsilon]$. With $u_C(t)\geq 0$, we will eventually have $x_C(t)>x_U(t)-[\varphi v_C(t)+\varepsilon]$ for all $t\in(t_1,t_f]$, which contradicts the safety condition (\ref{appeq:vehicle_dynamics}). Therefore, the safety constraint will never be activated for all $t\in[t_0,t_f]$ in this case.

Observe that in the analysis of Case 1, we apply $u_C(t)=\sqrt{2\alpha_t}$ to check all constraints and have not determined the value of $\alpha_t$. Therefore, if $\sqrt{2\alpha_t}>u_{C,\max}$ and the acceleration constraint is activated, we apply $u_{C,\max}$ instead. Then, the difference between $v_C(t_f)$ and $v_{flow}$ will be larger, so the speed and safety constraints will not be activated.\\
\\
\textbf{Case 2:} $x_C(t_f)\leq \bar{x}_C(t_f)\leq x_U(t_f)-\delta_C$

If $x_C(t_f)\leq \bar{x}_C(t_f)$, then $u_C^*(t)\leq 0$ and $u_C(t)=\sqrt{2\alpha_t}$ will be infeasible in this case. With the same analysis as in Case 1, $v_C(t)$ decelerates from $v_0$ to $v_{flow}$, hence $v_C(t)$ will not reach $v_{C,\min}$ since $v_{flow}$ satisfies $v_{flow}\geq v_{C,\min}$. Moreover, the safety constraint will not be activated for a reason similar to the one in Case 1.\\
\\
\textbf{Case 3:} $x_C(t_f)\leq x_U(t_f)-\delta_C \leq \bar{x}_C(t_f)$

For $x_C(t_f)\leq x_U(t_f)-\delta_C \leq \bar{x}_C(t_f)$, then $u_C^*(t)=-\sqrt{2\alpha_t}$. We can check if the control constraint will be activated for all $t\in[t_0,t_f]$. Thus, we only consider the speed and safety constraints next.  

Firstly, suppose only the speed constraint is active. Proceeding as in Case 1, the maximum speed constraint will not be activated. We only need to check if the minimum speed constraint is activated or not. Assume CAV $C$ reaches $v_{\min}$ at $t_1$, then $t_1 = t_0 + \dfrac{v_{\min}-v_C(t_0)}{-\sqrt{2\alpha_t}}$. However, after CAV $C$ enters the minimum speed-constrained arc, there is no hard terminal constraint to let $C$ exit this arc. 

Secondly, suppose only the safety constraint is activated at time $t_2\in[t_0,t_f)$. We then have $x_C(t_2)=x_U(t_2)-[\varphi v_C(t_2)+\varepsilon]$. To activate the safety constraint, we must have $v_C(t_2^-)>v_U$ and $x_C(t_2^-)<x_U(t_2^-)-[\varphi v_C(t_2)+\varepsilon]$. To guarantee safety, $v_C(t_2^+)$ should satisfy $v_C(t_2^+)\leq v_U$, otherwise, the safety constraint will be violated by the continuity of $x_C(t)$ and $v_C(t)$. Therefore, we have $v_C(t_2)=v_U$, which is equivalent to $v_0+u_C(t_2-t_0)=v_U$ and $t_2$ can be calculated as $t_2=t_0+\dfrac{v_U-v_0}{u_C}$. Since $u_C(t)=-\sqrt{2\alpha_t}<0$ for all $t\in[t_0,t_2]$ and $u_C(t)\leq 0$ for all $t\in[t_0,t_f]$, then $v_C(t)<v_U$ for all $t\in(t_2,t_f]$ if $t_2\leq t_f$, and the constraint will not be activated again in $t\in(t_2,t_f]$. If $t_2\geq t_f$, then the safety constraint will be inactive for all $t\in[t_0,t_f]$. However, it is impossible to get $t_2\geq t_f$, since $v_U<v_{flow}$, otherwise there is no reason for CAV $C$ to change its lane.

Lastly, if the two constraints are active, we need to figure out the activation order. If the minimum speed constraint is activated first with $v_{\min}<v_U$, the safety constraint will not be activated for all $t\in (t_1, t_f]$. Since the safety constraint is only activated instantaneously at $t_2\leq t_f$, the two constraints being active is equivalent to only the minimum speed constraint being active. 

From the analysis above, we have proved that the safety constraint may only be activated instantaneously at $t_2 \leq t_f$ for $v_U < v_{flow}$, which can be omitted, and then enter the minimum speed constrained arc. For $t\in(t_1,t_f]$, there are two possible trajectories for CAV $C$. One is to maintain $v_{\min}$, then $t_f=t_1$ for optimality (otherwise, the time component in (\ref{appeq:vehicle_dynamics_rewritten}) increases while the other two remain fixed).
Another trajectory is to accelerate again at $t_3<t_f$. If the safety constraint is inactive for all $t\in(t_3,t_f]$, i.e., $x_U(t)-x_C(t)>\delta_C(v_C(t)),\forall t\in(t_3,t_f],$ then the problem becomes unconstrained again, which is the same as Case 1. However, this trajectory will not be optimal because it contains unnecessary deceleration during $[t_0,t_1]$ without the safety constraint being activated for all $t\in[t_0,t_f]$ and causes increased energy consumption. Therefore, the optimal trajectory for CAV $C$ is to have the safety constraint activated at some $t_4\leq t_f$. If $t_4<t_f$, then we have $v_C(t)=v_U$ for all $t\in[t_4,t_f]$, otherwise the safety constraint will be violated or the trajectory will be again suboptimal. The total cost is increasing for $t$ from $t_4$ to $t_f$ since it causes additional time and energy without increasing speed. To achieve optimality, we conclude that $t_4=t_f$. 

Therefore, the inequality constraint $x_U(t)-x_C(t)= \delta_C(v_C(t))$ for all $t\in[t_0,t_f]$ is equivalent to $x_U(t_f)-x_C(t_f)= \delta_C(v_C(t_f))$ and the OCP for CAV $C$ can be rewritten as 
\begin{multline*}
    \label{appeq:cav_c_objective_simplified}   
    \min\limits_{t_f,u_C(t)} \alpha_t\left(t_f-t_0\right)+
    \frac{\alpha_{v}}{2}(v_C(t_f)-v_{flow})^2+
    \int_{t_0}^{t_f}\frac{1}{2}u_C^2(t) dt
\end{multline*}
\begin{equation}
    \begin{matrix}
        \dot{x}_C(t) = v_C(t), \;
    \dot{v}_C(t) = u_C(t), \; \; 
        \label{appeq:vehicle_dynamics_rewritten} \\
    u_{C_{\min}}\leq u_C(t)\leq u_{C_{\max}}, \; \; \forall t\in\lbrack t_{0},t_{f}\rbrack
        \\
        v_{C_{\min}}\leq v_C(t)\leq v_{C_{\max}}, \; \; \forall t\in\lbrack t_{0},t_{f}\rbrack,\\
        x_U(t_f)-x_C(t_f)= \delta_C(v_C(t_f)),\\
        0\leq t_f\leq T_{\max}
     \end{matrix}
\end{equation}
where the difference between (\ref{appeq:vehicle_dynamics}) and (\ref{appeq:vehicle_dynamics_rewritten}) is that the safety constraint for CAV $C$ with respect to vehicle $U$ becomes a strict equality constraint at the terminal time $t_f$. The reason is that we have proved that the safety constraint may be activated instantaneously at $t_2$ and will never be activated again after $t_2$ so that the safety constraint for all $t\in[t_0,t_f)$ becomes redundant. Moreover, equality holds to save any unnecessary deceleration of CAV $C$. For this revised OCP, its Hamiltonian becomes
\begin{multline}
    \label{eq:hamiltonian_re}
    H(\mathbf{x_C},\mathbf{\lambda_C},u_C)=
    \frac{1}{2}u_C^2(t)+\alpha_t+\lambda_C^x(t)v_C(t)+\lambda_C^v(t)u_C(t)\\
    +\mu_1(u_{C,\min}-u_C(t))+\mu_2(u_C(t)-u_{C,\max})\\
    +\mu_3(v_{C,\min}-v_C(t))+\mu_4(v_C(t)-v_{C,\max}),
\end{multline}
the costate boundary conditions become 
\begin{align*}
       &\lambda_C^x(t_f)=(\dfrac{\partial\Phi}{\partial x_C})|_{t=t_f}=a,\\
       &\lambda_C^v(t_f)=(\dfrac{\partial\Phi}{\partial v_C})|_{t=t_f}=\alpha_v[v_C(t_f)-v_{flow}],
\end{align*}
$a$ is a constant, and the Euler-Lagrange equations are given as 
\begin{align}
\label{appeq:eular_lagrange_re}
\nonumber\dot{\lambda}_C^x&=-\dfrac{\partial H}{\partial x_C}=0,\\
\dot{\lambda}_C^v&=-\dfrac{\partial H}{\partial v_C}=-\lambda_C^x+\mu_3-\mu_4.
\end{align}

\emph{(1) No Constraint Active.} 
If no constraint is active for all $t\in[t_0,t_f]$, then all Lagrange multipliers $\mu_1,\mu_2,\mu_3,\mu_4$ are 0. Thus, (\ref{appeq:eular_lagrange_re}) gives $\dot{\lambda}_C^x=0$, $\dot{\lambda}_C^v=-\lambda_C^x$, i.e., $\lambda_C^x=a,\lambda_C^v=-(at+b)$, where $a,b$ are integration constants. With the same necessary condition for optimality in (\ref{appeq:optimality_condition}), we have 
\begin{equation}
\label{appeq:optimal_condition_case3}
u_C(t)+\lambda_C^v(t)=0,
\end{equation}
which implies $u_C(t)=-\lambda_C^v(t)=at+b$. Hence, the speed and position trajectories of CAV $C$ will be expressed as  $v_C(t)=\frac{1}{2}at^2+bt+c$ and $x_C(t)=\frac{1}{6}at^3+\frac{1}{2}bt^2+ct+d,$ where $a,b,c,d$ are integration constants. In addition, the transversality condition (\ref{eq:transversality}) gives the following relationship:
\begin{equation}
    \label{appeq:transversality_relationship_no_constraint_active}
    \dfrac{1}{2}u_C^2(t_f)+\alpha_t+\lambda_C^x(t_f)v_C(t_f)+\lambda_C^v(t_f)u_C(t_f)=0.
\end{equation}
Combining all the boundary conditions above and the initial conditions of $t_0,v_C^0,v_U,v_{flow},\alpha_t,\alpha_v,x_C(t_0),x_U(t_0)$, the following six equations hold:
\begin{align}
\nonumber
&at_f+b=\alpha_v(v_{flow}-v_C(t_f)),\\
\nonumber
&v_C(t_f)=\frac{1}{2}at_f^2+bt_f+c,\\
\nonumber
&v_C(t_0)=\frac{1}{2}at_0^2+bt_0+c,\\
\nonumber
&\alpha_t+av_C(t_f)=\frac{1}{2}(at_f+b)^2,\\
\nonumber
&x_C(t_0)=\frac{1}{6}at_0^3+\frac{1}{2}bt_0^2+ct_0+d,\\
\label{appeq:unconstrained_equations}
&x_U(t_f)-\delta_C(v_C(t_f))=\frac{1}{6}at_f^3+\frac{1}{2}bt_f^2+ct_f+d.
\end{align}
Thus, the unknown variables $a,b,c,d,t_f$ can be obtained by solving (\ref{appeq:unconstrained_equations}). It is also easy to show that $a$ has to be non-negative, and $b$ has to be non-positive. 
If both $a,b$ are positive, then CAV $C$ will never decelerate, which violates the terminal conditions for $x_C(t_f)\leq \bar{x}_C(t_f)$. If both $a,b$ are negative, then CAV $C$ will keep decelerating, which violates the optimality in the unconstrained case. If $a$ is negative, $b$ is positive, this means that CAV $C$ will accelerate over $[t_0,t_1]$ where $t_1=-\frac{b}{a}$, then decelerate. If $t_1\geq t_f$, which means CAV $C$ keeps accelerating for all $t\in[t_0,t_f]$, then the terminal constraint $x_C(t_f)\leq \bar{x}_C(t_f)$ will be violated. If $t_1\leq t_f$, which means CAV $C$ accelerates in $[t_0,t_1]$, then decelerates in $[t_1,t_f]$, which violates the optimality in the unconstrained case.  
\\
\emph{(2) Safety Constraint Active Only.}
In this case, we can show that the safety constraint will be activated at most once instantaneously during $[t_0,t_f).$
First, note that with the initial condition $v_C(t_0)<v_{flow}$ and the terminal position $x_C(t_f)\leq x_U(t_f)-\delta_C \leq \bar{x}_C(t_f)$, CAV $C$ cannot keep accelerating over $[t_0,t_f]$, otherwise we will get Case 1. If the safety constraint is activated at $t_1$, then $v_C(t_1)=v_U$; otherwise, if $v_C(t_1)>v_U$, then from the continuity of velocity, the safety constraint will be violated at $t_1^+$, whereas if $v_C(t_1)<v_U$, then the safety constraint will not be activated at $t_1$. Considering the value of $t_1$, if $t_1\geq t_f$, then it follows that indeed the safety constraint will be activated at most once. On the other hand,
if $t_1<t_f$, there are two possible trajectories for CAV $C$ in $t\in[t_1,t_f]$. One is to travel at constant speed such that the safety constraint is activated for all
    $t\in[t_1,t_f]$ with   $v_C(t)=v_U$. However, to achieve optimality, $t_1$ should be exactly equal to $t_f$ because the objective function in (\ref{appeq:vehicle_dynamics_rewritten})
    is monotonically increasing over $[t_1,t_f]$ with the same terminal speed cost. Another trajectory is to exit the safety-constrained arc at $t_2<t_f$. However, to guarantee safety, $C$ still has to decelerate at $t_2$, then accelerate at some $t_3$ such that $t_2<t_3<t_f$ to satisfy the terminal position constraint $x_U(t_f)-x_C(t_f)= \delta_C(v_C(t_f)).$ However, decreasing the length of the safety-constrained arc will decrease the objective, hence the optimal solution is to have $t_1=t_2$. 

Therefore, having shown that the safety constraint will be activated at most once instantaneously during $[t_0,t_f)$, we now only need to consider whether the speed and control constraints are active as discussed in the remaining cases below.
\\
\emph{(3) Control Constraint Active Only.}
The analysis in \emph{(1)} indicates that the acceleration is non-decreasing, so the minimum control constraint will not be activated if it is inactive at the initial time. Firstly, we assume only the maximum control constraint is activated. Suppose $u(t)$ reaches $u_{\max}$ at time $t_1$. Then $u_C(t)=at+b$ for $t\in[t_0,t_1]$, $u_C(t)=u_{\max}$ for $t\in[t_1,t_f]$ and $t_1$ can be computed by $at_1+b=u_{\max}$ with the obtained coefficients $a,b$ from (\ref{appeq:unconstrained_equations}). Moreover, $x_C(t_1),v_C(t_1)$ can also be obtained and we can express the terminal speed as
\begin{gather}\label{appeq:control_terminal_speed}
v_C(t_f)=v_C(t_1)+u_{\max}(t_f-t_1),
\end{gather}
and the terminal position 
\begin{align}\label{appeq:control_terminal_position}
\nonumber
v_U(t_f-t_0)&-\delta(v_C(t_f))\\
&=x_C(t_1)+v_C(t_1)(t_f-t_1)+\frac{1}{2}u_{\max}(t_f-t_1)^2
\end{align}
to solve for the terminal time $t_f$.

If only the minimum control constraint is activated, since the acceleration is non-decreasing, the entry point of the constrained arc is $t_0$, and suppose $t_2$ is the exit point of the constrained arc. Similarly, we have the following six equations:
\begin{align}\label{appeq:control_equations}
\nonumber
&at_2+b=u_{\min},\\
\nonumber
&at_f+b=\alpha_v [v_{flow}-v_C(t_f)],\\
\nonumber
&\frac{1}{2}at_2^2+bt_2+c=v_{\min},\\
\nonumber
&\frac{1}{6}at_2^3+\frac{1}{2}bt_2^2+ct_2+d=\\
\nonumber 
& \; \; \; \; \; \; \; \; \; \;  \; \; \; x_C(t_0)+v_C(t_0)(t_2-t_0)+\frac{1}{2}u_{\min}(t_2-t_0)^2,\\
\nonumber
&\frac{1}{6}at_f^3+\frac{1}{2}bt_f^2+ct_f+d=x_U(t_f)-\delta(v_C(t_f)),\\
&\frac{1}{2}(at_f+b)^2+\alpha_t+av_C(t_f)-(at_f+b)^2=0,
\end{align}
to solve for $a,b,c,d,t_2,t_f$.

Finally, if the minimum and maximum constraints are both activated, suppose the exit point for the minimum constraint is $\tau_1$, and the starting point for the maximum constraint is $\mu_2$. Then $u(t)=u_{\min},t\in[t_0,\tau_1)$, $u(t)=at+b,t\in[\tau_1,\tau_2]$ and $u(t)=u_{\max},t\in(\tau_2,t_f]$. Therefore, $\tau_1$ is the same as $t_2$ in the above case, and we now calculate $\tau_2$ and $t_f$ by using $a\tau_2+b=u_{\max}$  with
\begin{align}
\nonumber
x_C(\tau_2)+v_C(&\tau_2)(t_f-\tau_2)\\
&+\frac{1}{2}u_{\max}(t_f-\tau_2)^2=x_U(t_f)-\delta(v_C(t_f))
\end{align}
\begin{align*}
    &v_C(\tau_2)=\frac{1}{2}a\tau_2^2+b\tau_2+c,\\
    &x_C(\tau_2)=\frac{1}{6}at_2^3+\frac{1}{2}bt_2^2+ct_2+d
\end{align*}
so that we can solve for $\tau_2,t_f$, where $a,b,c,d$ are obtained from the equations (\ref{appeq:control_equations}).
\\
\emph{(4) Speed Constraint Active Only.}
From the terminal position, we know that CAV $C$ includes both a deceleration and acceleration trajectory segment. Since the safety constraint will never be activated in $[t_0,t_f)$ and we have $v_{flow}<v_{\max}$, the maneuver will be completed before the maximum constraint is activated. Hence, we only need to check whether the minimum speed $v_{\min}$ is reached.  Suppose $t_1$ is the starting point of such a constrained arc and $t_2$ is the exit point. Then, $u(t)=at+b$ for $t\in[t_0,t_1),$ and $u(t)=0$ for $t\in[t_1,t_2]$. We can compute $t_1$ through $at_1+b=0$ and further express the velocity and position of CAV C at $t_1$ as
\begin{align}\label{appeq:minspeed_entry}
\nonumber
v_C(t_1)&=\frac{1}{2}at_1^2+bt_1+c,\\
x_C(t_1)&=\frac{1}{6}at_1^3+\frac{1}{2}bt_1^2+ct_1+d,
\end{align} 
where the coefficients $a,b,c,d$ are obtained by (\ref{appeq:unconstrained_equations}). Hence, we can solve the following six equations 
\begin{align}\label{appeq:minspeed_equations}
\nonumber
&a_1t_2+b_1=0,\\
\nonumber
&\frac{1}{2}a_1t_2^2+b_1t_2+c_1=v_{\min},\\
\nonumber
&a_1t_f+b_1=\alpha_v(v_{flow}-v_C(t_f)),\\
\nonumber
&\frac{1}{6}a_1t_2^3+\frac{1}{2}b_1t_2^2+c_1t_2+d_1=x_C(t_1)+v_{\min}(t_2-t_1),\\
\nonumber
&\frac{1}{6}a_1t_f^3+\frac{1}{2}b_1t_f^2+c_1t_f+d_1= x_U(t_f)-\delta(v_C(t_f)),\\
&\frac{1}{2}(a_1t_f+b_1)^2+\alpha_t+a_1v_C(t_f)-(a_1t_f+b_1)^2=0
\end{align} 
to obtain $a_1,b_1,c_1,d_1,t_2,t_f$.
\\
\emph{(5) Control and Speed Constraints Active.}
Based on the speed constraint analysis above, the maximum speed will not be reached during $[t_0,t_f]$. Therefore, there are only three cases that could occur if we have control and speed constraints active, i.e., (\romannumeral1) $u_{\min}$, $v_{\min}$ active, (\romannumeral2) $u_{\max}$, $v_{\min}$ active, (\romannumeral3) $u_{\min}$, $v_{\min}$ and $u_{\max}$ active.\\
(\romannumeral1) $u_{\min}$, $v_{\min}$ active:
First, we need to check which constraint will be activated first. If $v_{\min}$ is reached, this indicates that the acceleration decreases to 0, and the acceleration has to be positive to exit the $v_{\min}$ constrained arc. Hence, $u_{\min}$ is reached first at $t_1>t_0$, then reaches $v_{\min}$ at $t_2>t_1$.  Notice that for $t\in[t_0,t_1)$, the control constraint has the same performance as in \emph{(3)}, so we use (\ref{appeq:control_equations}) to get the CAV $C$ trajectory for $t\in[t_0,t_1]$ directly.

At time $t_2$, we have the interior constraint $N(v_C(t))=v_{\min}-v_C(t_2)=0$, and 
\begin{align}\label{appeq:interior_constraint}
\nonumber
\lambda_C^T(t_2^-)&=\lambda_C^T(t_2^+)+\pi \dfrac{\partial N}{\partial \textbf{x}}|_{t=t_2},\\
H(t_2^-)&=H(t_2^+)-\pi \dfrac{\partial N}{\partial t}|_{t=t_2},
\end{align}
where $\pi$ is a Lagrangian multiplier. Based on (\ref{appeq:interior_constraint}), we have $\lambda_C^x(t_2^-)=\lambda_C^x(t_2^+)$, $\lambda_C^v(t_2^-)=\lambda_C^v(t_2^+)-\pi$, and $H(t_2^-)=H(t_2^+)$. Hence, we have 
\begin{multline}
    \label{appeq:interior_transversality}
    \frac{1}{2}u_C^2(t_2^-)+\alpha_t+\lambda_C^x(t_2^-)v_C(t_2^-)+\lambda_C^v(t_2^-)u_C(t_2^-)\\
    =\frac{1}{2}u_C^2(t_2^+)+\alpha_t+\lambda_C^x(t_2^+)v_C(t_2^+)+
    \lambda_C^v(t_2^+)u_C(t_2^+).
\end{multline}   
Since the state variables are continuous, we have $v_C(t_2^-)=v_C(t_2^+)$, $x_C(t_2^-)=x_C(t_2^+)$ and $u_C(t_2^+)=0$. Moreover, the optimality condition gives $u_C(t_2^-)=-\lambda_C^v(t_2^-)$. Thus, we can calculate $u_C(t_2^-)=0$ according to (\ref{appeq:interior_transversality}), i.e., $u_C(t_2)=0$. Letting the exit point be $t_3$, combining the optimality and transversality conditions, we have the following six equations
\begin{align}
\nonumber
&\frac{1}{2}a_1t_3^2+b_1t_3+c_1=v_{\min},\\
\nonumber
&\frac{1}{6}a_1t_3^3+\frac{1}{2}b_1t_3^2+c_1t_3+d_1=x_C(t_2)+v_{\min}(t_3-t_2),\\
\nonumber
&a_1t_3+b_1=0,\\
\nonumber
&\frac{1}{6}a_1t_f^3+\frac{1}{2}b_1t_f^2+c_1t_f+d_1= x_U(t_f)-\delta(v_C(t_f)),\\
\nonumber
&a_1t_f+b_1=\gamma(v_d-v_C(t_f)),\\
&\frac{1}{2}(a_1t_f+b_1)^2+\beta+a_1v_C(t_f)-(a_1t_f+b_1)^2=0
\end{align}
to solve for $a_1,b_1,c_1,d_1,t_3,t_f$. Note that using the coefficients $a,b,c,d$ in (\ref{appeq:control_equations}) and the known $t_1$, we can similarly calculate $t_2$ from $at_2+b=0$, and $v_C(t_2),x_C(t_2)$.\\
(\romannumeral2) \emph{$u_{\max}$, $v_{\min}$ active:} 
In this case, the optimality of the solution determines that $v_{\min}$ has to be reached before $u_{\max}$, otherwise, $v_{\min}$ will not be reached or the solution will not be optimal. For the minimum speed constraint being activated first, suppose $t_1$ is the entry point of the minimum velocity-constrained arc, $t_2$ is the corresponding exit point, and $t_3$ is the entry point of the acceleration constraint. For $t\in[t_0,t_1)$, the trajectories of CAV $C$ have the same performance as in \emph(4), i.e., $u_C(t)=at+b,t\in[t_0,t_1)$, $u_C(t)=0,t\in[t_1,t_2)$, and $u_C(t)=a_1t+b_1,t\in[t_2,t_3),$ where $t_1,t_2,a,b,a_1,b_1$ can be calculated from (\ref{appeq:minspeed_entry}) and (\ref{appeq:minspeed_equations}). In addition, $t_3$ can be computed from $a_1t_3+b_1=u_{\max}$, and we can further obtain $v_C(t_3),x_C(t_3)$. 

For the terminal position constraint, the states at the terminal time $t_f$ should satisfy the following equation:
\begin{align*}
x_C(t_3)+v_C(t_3)(t_f&-t_3)+\frac{1}{2}u_{\max}(t_f-t_3)^2\\
&=x_U(t_0)+v_U(t_f-t_0)-\phi v_C(t_f)-\delta,
\end{align*}
which provides the solution for $t_f$.\\
(\romannumeral3) $u_{\min}$, $v_{\min}$ and $u_{\max}$ active:
regarding the activation order of the three constraints, $u_{\min}$, $v_{\min}$, and $u_{\max}$,
proceeding as in the previous cases, $u_{\min}$ will be activated first to reach $v_{\min}$, then CAV $C$ accelerates from $v_{\min}$ to $v_{flow}$, and then $u_{\max}$ is activated. We can directly combine the results of cases (\romannumeral1) and (\romannumeral2) to get the complete trajectory for CAV $C$ in this case. 

Lastly, a similar analysis can be performed for the solutions for CAVs $i$ and $i+1$, thus we omit the details for the Hamiltonian analysis corresponding to  \eqref{eq:cav_i_problem}, \eqref{eq:cav_i_1_problem}, and \eqref{eq:cav_c_relaxedTime}.